\documentclass[11pt, a4paper, reqno]{amsart}

\usepackage[utf8]{inputenc}
\usepackage[T1]{fontenc}
\usepackage{amsmath, mathtools, amsthm}
\usepackage{mathrsfs}
\usepackage{amssymb,url,xspace}
\usepackage[dvipsnames]{xcolor}
\usepackage{cases}
\usepackage[hidelinks]{hyperref}
\usepackage{cleveref}
\usepackage{tikz}
\usepackage{verbatim}
\usetikzlibrary{matrix}
\usepackage{xfrac}
\usepackage[outercaption]{sidecap} 
\usepackage[percent]{overpic}
\usepackage[numbers]{natbib}

\definecolor{dukeblue}{rgb}{0.0, 0.0, 0.61}
\definecolor{darkcandyapplered}{rgb}{0.64, 0.0, 0.0}
\definecolor{firebrick}{rgb}{0.7, 0.13, 0.13}
\definecolor{brandeisblue}{rgb}{0.12, 0.46, 0.7}
\definecolor{goldenrod}{rgb}{0.85, 0.65, 0.13}
\definecolor{tabpurple}{rgb}{0.57, 0.4, 0.73}

\hypersetup{
	colorlinks,
	citecolor=dukeblue,
	filecolor=black,
	linkcolor=dukeblue,
	urlcolor=black
}
\usepackage{empheq}

\usepackage[svgnames,pdf]{pstricks}
\usepackage{anysize}
\marginsize{3.35cm}{3.35cm}{3.25cm}{3.25cm}

\newcommand{\diff}{\, \mathrm{d}}

\newcommand{\R}{\mathbb{R}}

\def\Lip{\mathrm{Lip}}
\def\loss{\mathrm{loss}}

\def\*#1{\mathbf{#1}}

\theoremstyle{plain}
\numberwithin{equation}{section}
\newtheorem{proposition}{Proposition}[section]
\newtheorem{theorem}{Theorem}[section]
\newtheorem{lemma}{Lemma}[section]
\newtheorem{cor}{Corollary}[section]

\newtheorem{assumption}{Assumption}
\newtheorem{definition}{Definition}[section]
\newtheorem{remark}{Remark}
\theoremstyle{definition}

	\title 
	[Sparsity in long-time control of neural ODEs]{Sparsity in long-time control of neural ODEs}

	\author[Carlos Esteve-Yagüe]{Carlos Esteve-Yagüe}	
	\author[Borjan Geshkovski]{Borjan Geshkovski}	
	\date{\today}

\begin{document}
	
		\begin{abstract}
			We consider the neural ODE and optimal control perspective of supervised learning, with $\ell^1$-control penalties, where rather than only minimizing a final cost (the \emph{empirical risk}) for the state, we integrate this cost over the entire time horizon. We prove that any optimal control (for this cost) vanishes beyond some positive stopping time. When seen in the discrete-time context, this result entails an \emph{ordered} sparsity pattern for the parameters of the associated residual neural network: ordered in the sense that these parameters are all $0$ beyond a certain layer.
			Furthermore, we provide a polynomial stability estimate for the empirical risk with respect to the time horizon.
			This can be seen as a \emph{turnpike property}, for nonsmooth dynamics and functionals with $\ell^1$-penalties, and without any smallness assumptions on the data, both of which are new in the literature. 
		\end{abstract}
		
	\maketitle	
	\setcounter{tocdepth}{1}
	\tableofcontents
	
	{\small
		{\bf Keywords.} Deep Learning; Neural ODEs; Supervised Learning; Sparsity; Optimal control; \indent Turnpike property, Stabilization.}

	{\small\href{https://mathscinet.ams.org/msc/msc2010.html}{{\bf 	\color{dukeblue}{AMS Subject Classification}}}}. 49J15; 49M15; 49J20; 49K20; 93C20; 49N05.

\section{Introduction}

\subsection{Motivation} Sparsity is a highly desirable property in many machine learning and optimization tasks due to the inherent reduction of computational complexity.
Typically induced by $\ell^1$ penalties/regularizations, it has been used extensively for simplifying machine learning tasks by selecting, in an automatized manner, a strict subset of the available features to be used.  
This is exemplified by the well-known Lasso (least absolute shrinkage and selection operator, \citep{santosa, tibshirani1996regression}), which consists in minimizing a least squares cost function and an $\ell^1$ parameter penalty for an affine parametric model $y=wx+b$. As the $\ell^1$ penalty enforces a subset of the optimizable parameters $(w,b)$ to become zero, the associated features may be discarded safely.

With such insights in mind, in this work we analyze supervised learning problems viewed from the lens of optimal control and neural ODEs, and demonstrate the appearance of sparsity patterns for global minimizers in the context of $\ell^1$ control penalties. Rather than typical sparsity in which, at a given time $t$, all but few of the components of a control $u(t)\in\mathbb{R}^{d_u}$ are zero, we shall demonstrate a \emph{ordered} or \emph{temporal} sparsity: an optimal control $u(t)$ concentrates all its amplitude within a subinterval $[0,T^*]$ (wherein it may very well be additionally sparse
), and vanishes beyond time $t\geqslant T^*$. (See \Cref{thm: L1.regu})

We motivate our setting and main result in what follows, and refer the reader to \Cref{sec: outline} for a roadmap of the paper. 

\subsection{Supervised learning}
To put the above discussion into context, we recall that \emph{supervised learning} addresses the problem of predicting from labeled data, which consists in approximating an unknown function $f:\mathcal{X}\to\mathcal{Y}$ from known samples 
\begin{equation*}
\left\{x^{(i)}, y^{(i)}\right\}_{i\in[n]}\subset\mathcal{X}\times\mathcal{Y}.
\end{equation*}
Here and henceforth, $[n]:=\{1,\ldots,n\}$ and $\mathcal{X}\subset\mathbb{R}^d$. 
	Depending on the nature of the label space $\mathcal{Y}$, one distinguishes two types of supervised learning tasks: \emph{classification}, when labels take values in a finite set of $m\geqslant2$ classes, e.g. $\mathcal{Y}=[m]$,  and \emph{regression}, when the labels take continuous values in $\mathcal{Y}\subset \R^m$ with $m\geqslant1$. 
	To solve a supervised learning problem, one seeks to construct a map 
	$f_{\text{approx}}:\mathcal{X}\to\mathcal{P}(\mathcal{Y})$,
which, desirably, is such that for any $x\in \mathcal{X}$ and for any Borel measurable $A\subset \mathcal{Y}$, $f_{\text{approx}}(x)(A)\simeq 1$ whenever $f(x)\in A$, and $f_{\text{approx}}(x)(A)\simeq 0$ whenever $f(x)\not\in A$; here, $\mathcal{P}(\mathcal{Y})$ denotes the space of probability measures on $\mathcal{Y}$.
In other words, one looks for a map $f_{\text{approx}}$ which approximates the map $x\to\delta_{f(x)}$ where $\delta_z$ denotes the Dirac measure centered at $z$. 
Ultimately, this translates to simultaneously interpolating the above dataset through $f_{\text{approx}}$, whilst ensuring generalization/extrapolation, namely reliable prediction on points in $\mathcal{X}$ which are outside of said dataset (\citep{zhang2016understanding}).

\subsection{An optimal control perspective}
There are various ways in which one can construct such an approximation $f_{\text{approx}}$, with different degrees of empirical and theoretical guarantees.
In this paper,  following a recent trend started with the works \citep{weinan2017proposal, haber2017stable, chen2018neural}, we shall focus on parametrizing $f_{\text{approx}}$ by the flow of neural ODEs, such as
	\begin{equation} \label{eq: 1.2}
	\begin{dcases}
	\dot{\*x}_i(t) = w(t)\sigma(\*x_i(t))+b(t) & \text{ for } t \in (0, T), \\
	\*x_i(0) = x^{(i)}\in\mathbb{R}^d,
	\end{dcases}
	\end{equation} 
	for $i\in[n]$ and $T>0$, with $\sigma$ being a scalar, globally Lipschitz function defined componentwise in \eqref{eq: 1.2}. The matrix $w(t)\in\mathbb{R}^{d\times d}$ and vector $b(t)\in\mathbb{R}^d$ play the role of controls (called \emph{parameters} in machine learning jargon), which in practice are found by solving an empirical risk minimization problem of the form
	\begin{equation} \label{empiric loss}
	\inf_{\substack{u=(w,b)\in\mathfrak{U}\\ \*x_i \text{ solves } \eqref{eq: 1.2}}} \underbrace{\frac{1}{n} \sum_{i=1}^n \loss\left(P\*x_i(T), y^{(i)}\right)}_{:=\mathscr{E}(\*x(T))} + \int_0^T \|u(t)\|_1\diff t.
	\end{equation}
	Here, $\mathfrak{U}$ is an appropriate Banach subspace of $L^1(0,T;\mathbb{R}^{d_u})$, $P: \R^d\to\R^m$ is an affine map which we suppose to be given\footnote{In practice, $P$ is either an optimizable variable, or its coefficients may be chosen at random. While we fix $P$ for technical purposes, our numerical experiments indicate that the results presented in what follows persist when $P$ is optimized as well.}, 
	and which serves to match the states $\*x_i(T)$ with the labels $y^{(i)}$ (typically of different dimensions), while 
	\begin{equation*}
	\loss(\cdot,\cdot): \R^m \times \mathcal{Y}\to\R_+
	\end{equation*}
	is such that $x\mapsto\loss(x,y)$ is continuous for all $y\in\mathcal{Y}$, $\loss(x,y)\neq 0$ whenever $\mu(x)\neq\delta_y$, and $\loss(x , y)\to0$ when $\mu({x})\to\delta_y$ in an appropriate sense of measures (e.g., for some Wasserstein distance, or for the Kullback-Leibler divergence). 
	A prototypical example is given by the square of the euclidean distance (\emph{least squares error}).
	But more tailored loss functions may be used, including positive and non-coercive ones, such as the \emph{cross-entropy} loss commonly used for classification tasks
	\begin{equation} \label{eq: cross.entropy}
	\loss\big(x, y\big) := -\log\left(\frac{e^{x_{y}}}{\sum_{j=1}^m e^{x_j}}\right) \hspace{1cm} \text{ for } x\in\mathbb{R}^m,\, y\in[m].
	\end{equation}
	 Once a solution $u=(w,b)$ to \eqref{empiric loss} is found, one may construct the approximation $f_{\text{approx}}$ by setting $f_{\text{approx}}(x)=\mu(\*x(T))$ for $x\in\mathcal{X}\subset\mathbb{R}^d$, where $\*x(T)$ solves \eqref{eq: 1.2} with $\*x(0) = x$ and control $u$.
	 The choice of $\mu:\mathcal{X}\to\mathcal{P}(\mathcal{Y})$ depends on the loss function and task at hand; for the least squares error loss for instance, one sets $\mu(x) := \delta_{Px}$, while for the cross-entropy loss, one sets $\mu := \text{softmax}\circ P$, with $\text{softmax}(z)_\ell=\sfrac{e^{z_{\ell}}}{\sum_{j=1}^m e^{z_j}}$ for $\ell\in[m]$ and $z\in\mathbb{R}^m$, as in \eqref{eq: cross.entropy} (designating a smooth approximation of the \text{argmax}). 
	 
	 The above presentation thus leads one to note that, in the neural ODE setting, supervised learning is a particular optimal control problem, wherein one looks to find a single pair of controls $u=(w,b)$, which steer $n$ trajectories of a nonlinear ODE such as \eqref{eq: 1.2}, corresponding to $n$ different initial data, to $n$ different targets. 
	  	 
	 \subsection{The role of $T$} Let us motivate our reason for considering the neural ODE and optimal control interpretation of supervised learning. 
	 In practice, one typically considers some discrete-time analog of \eqref{eq: 1.2}, e.g. a forward Euler scheme of the form
	 \begin{equation} \label{eq: resnet}
	 \begin{dcases}
	 \*x^{k+1}_i = \*x^k_i + \bigtriangleup t \left(w^k\sigma\big(\*x_i^k\big) + b^k\right) &\text{ for } k \in \{0, \ldots, n_t-1\},\\
	 \*x^0_i = x^{(i)},
	 \end{dcases}
	 \end{equation}
	 for $i\in[n]$, where $n_t\geqslant2$ and $\bigtriangleup t=\sfrac{T}{n_t}$. The scheme \eqref{eq: resnet} is an example of a \emph{residual neural network} (ResNet), a popular neural network architecture introduced in \citep{he2016deep}. 
	 As shown in \citep{he2016deep}, such neural networks provide, empirically, remarkable interpolation \emph{and} extrapolation performance when $n_t$ is large (of the orders of hundreds). 
	  Here, $n_t$ is referred to as the \emph{depth} of the network \eqref{eq: resnet} and each time-step $k$ is called a \emph{layer}. 
	  However, the theory supporting these empirical results is not completely mature (\citep{zhang2016understanding}).
	 
	 We observe that when $\bigtriangleup t>0$ is fixed, the time horizon $T$ can be used to estimate the depth $n_t$. 
	 This warrants the study of the behavior of optimal control problems for neural ODEs when $T$ is increased.
	On another hand, for many problems in optimal control, tracking the control and the trajectory over the entire time interval yields quantitative stability estimates for both when $T$ is large enough. This is for instance the case in turnpike theory for linear quadratic (LQ) problems (\citep{geshkovski2022turnpike}). 
	This setup is further motivated by empirical studies in machine learning literature, where a penalty of the state over each layer has been seen to yield better larger margin predictors, and thus better generalization, for specific classification tasks (\cite{elsayed2018large}).
	Consequently, in this work, rather than \eqref{empiric loss}, we are led to consider
	 \begin{equation} \label{eq: 1.4}
	 \inf_{\substack{u=(w,b)\in\mathfrak{U}\\ \*x_i \text{ solves } \eqref{eq: 1.2}}} \int_0^T \mathscr{E}(\*x(t))\diff t + \int_0^T \|u(t)\|_1\diff t,
	 \end{equation}
	 where $\mathscr{E}$ is defined in \eqref{empiric loss}, and where we set $\*x(t)=\{\*x_i(t)\}_{i\in[n]}$.
	 Our goal in this work is to provide a rather complete picture of the behavior of solutions to \eqref{eq: 1.4} and \eqref{eq: 1.2} as functions of $T$. 
	 
	 \subsection{Our contributions} \label{sec: whatever}
\begin{figure}[h]
	 \begin{overpic}[scale=0.5]{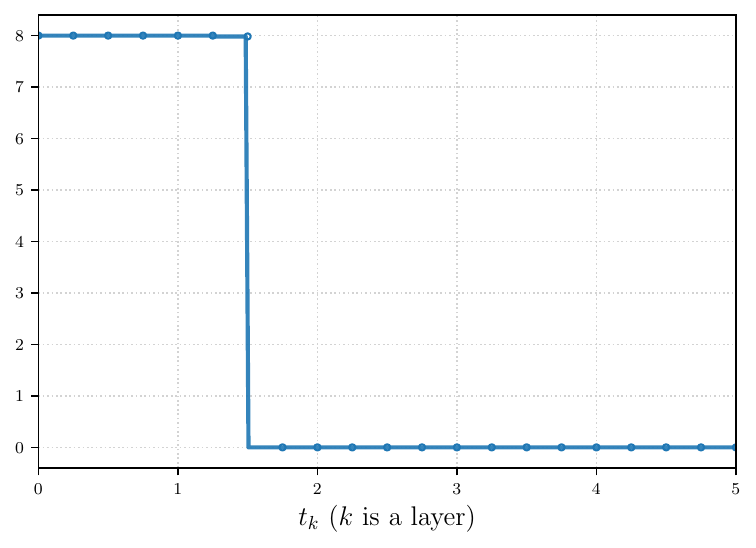}
\put (80, 17) {{\rotatebox{0}{{\color{brandeisblue}\tiny$\|u(t)\|_1$}}}}
\end{overpic}
\begin{overpic}[scale=0.5]{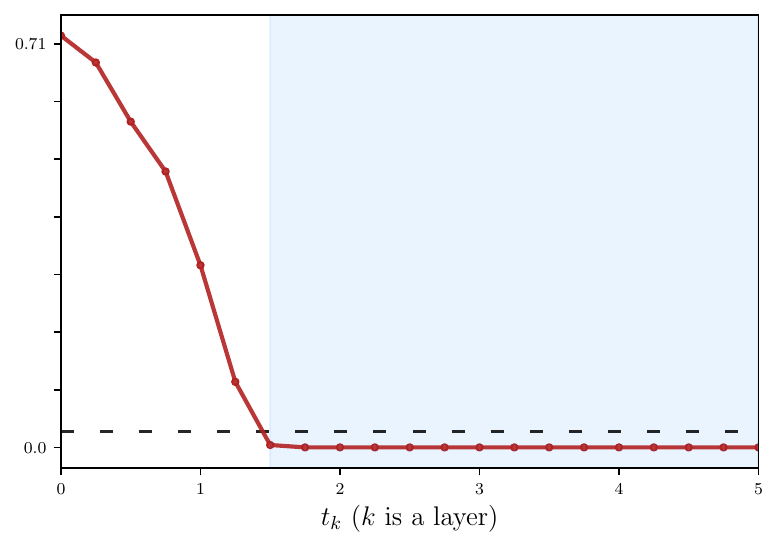}
\put (23.5, 48) {{\rotatebox{-62.5}{{\color{firebrick}\tiny$\mathscr{E}(\*x(t))$}}}}
\put (67.5, 18) {{\rotatebox{0}{\tiny$\mathcal{O}\left(\frac{1}{MT}+\frac{1}{T}\right)$}}}
\end{overpic}
\caption{(\emph{Left}) Optimal controls $u(t)$ solving \eqref{eq: 1.4}. (\emph{Right}) The empirical risk $\mathscr{E}(\*x(t))$ of the optimal states $\{\*x_i(t)\}_{i\in[n]}$. 
Both vanish beyond time $T^*=1.5$, which corresponds to $7$ layers.}
 \label{fig: fig1}
\end{figure}

	 We can illustrate our findings through numerical experiments\footnote{The \texttt{PyTorch} code may be found at \href{https://github.com/borjanG/dynamical.systems}{\textcolor{dukeblue}{\texttt{https://github.com/borjanG/dynamical.systems}}}.} before proceeding with theoretical setups and proofs. 
	 In \Cref{fig: fig1} (see \Cref{fig: fig3.5}--\Cref{fig4} for related illustrations), we depict a solution of \eqref{eq: 1.4} for a binary classification task ($\mathcal{Y}:= \{1,2\}$, with the data in \Cref{fig3}), with $\sigma\equiv\tanh$, using the cross-entropy loss defined in \eqref{eq: cross.entropy}, $T=5$, $\bigtriangleup t=\sfrac14$ (thus $20$ ResNet layers) with a midpoint scheme, and $n=3000$. We also impose the constraint $\|u(t)\|_1\leqslant M$ with $M=8$, to avoid concentration near $t=0$. (See \Cref{rem: constraint}.)
	 
{\color{black}
\begin{itemize}
\item The numerics show that optimal controls $u_T(t)=(w_T(t),b_T(t))$ concentrate within a subinterval $[0,T^*]$, and vanish beyond time $T^*$ (the \emph{ordered} sparsity pattern we had alluded to). The corresponding states $\{\*x_i(t)\}_{i\in[n]}$ are not only stationary for $t\geqslant T^*$, but actually in the regime in which $\mathscr{E}(\*x(t))$ is near $0$, as desired. 


\smallskip
\item In practical terms, the ordered sparsity and stability results could then be used to discard unnecessary layers in the corresponding residual neural network (ResNet), without removing relevant information. They also provide a quantitative estimate of the number of layers needed to fit the data, whilst keeping the controls of user-prescribed amplitude (thus possibly helping in generalization). These estimates ensure and indicate that the time horizon (or number of layers) ought not to be large at all for the error to reach $0$ (\Cref{fig: fig1}). 
\smallskip 

\item 
However, the presence of a minimal time $T^*$ would mean that we still need several layers – namely a large enough $T$ – before entering the stability regime, from which point on the empirical risk can be ensured to be small. This implies
a trade-off in how large $T$ should actually be. One should keep in mind that our numerical experiments are toy examples and do not convey possible difficulties encountered for various real-life datasets, which may be significantly more complex. (This complexity can partially be seen through our upper bounds in \Cref{thm: L1.regu}, see \Cref{rem: dep.data}.)
\smallskip

\item
All in all, in computing terms, the pointwise stability estimate further indicates that a hybrid, model predictive control (MPC)-type strategy is warranted for an optimal choice of the stopping time (see \citep{grune2019sensitivity, esteve2020large} for similar considerations). Our theoretical results provide further backbone for such ideas, which have been used in applied scenarios (\cite{goodfellow2016deep}).
\end{itemize}
}

In the subsequent section, we shall mathematically formalize these results (\Cref{thm: L1.regu}) and provide rigorous proofs ensuring their validity in a wide array of functional settings.
	 
\subsection{Outline} \label{sec: outline}
The remainder of this work is structured as follows. 
In \textbf{\Cref{sec: section.2}}, we provide the functional setting and our main result (\Cref{thm: L1.regu}), which corroborates the numerical experiment presented just above. Further numerical visualizations of the same experiment may also be found therein. The proof of \Cref{thm: L1.regu} may be found in \textbf{\Cref{sec: proofs}}. 
We conclude with a selection of open problems in \textbf{\Cref{sec: outlook}}.
	  

\section{Main result} \label{sec: section.2}

\subsection{Setup} 

We henceforth suppose we are given a dataset 
\begin{equation} \label{eq: dataset}
\left\{x^{(i)}, y^{(i)}\right\}_{i\in[n]}\subset\mathcal{X}\times\mathcal{Y}
\end{equation}
with $\mathcal{X}\subset\mathbb{R}^d$ and $x^{(i)}\neq x^{(j)}$ for $i\neq j$. The label space $\mathcal{Y}$ may either be a finite subset of $\mathbb{N}$, or a subset of $\mathbb{R}^m$. To have a more coherent presentation and simplify the technical details, we shall stack all of the trajectories $\*x_i(t)$ appearing in neural ODEs as \eqref{eq: 1.2}, in order, into one single vector $\*x(t)\in\mathbb{R}^{dn}$. 
Namely, we set 
\begin{equation*}
\*x(t):=
\begin{bmatrix}
\*x_1(t)\\
\vdots\\
\*x_n(t)
\end{bmatrix}\in\mathbb{R}^{d_x}, 
\hspace{1cm}
\*x^0:=
\begin{bmatrix}
x^{(1)}\\
\vdots\\
x^{(n)}
\end{bmatrix}
\in\mathbb{R}^{d_x}
\end{equation*}
for $i\in[n]$ and $t\geqslant0$, where $d_x:=dn$, and consider stacked neural ODEs in the general form 
\begin{equation} \label{dyn.general}
\begin{dcases}
\dot{\*x}(t) = \*f(\*x(t), u(t)) &\text{ for } t \in (0,T),\\
\*x(0) = \*x^0,
\end{dcases}
\end{equation}
where $u(t):=(w(t),b(t))\in\mathbb{R}^{d^2+d}$.
As presented in \eqref{eq: 1.2}, for the stacked system the nonlinearity $\*f: \R^{d_x}\times \R^{d_u}\to\R^{d_x}$ may take the form
\begin{equation}\label{dyn.sigma.inside intro}
\*f(\*x,u)  = \begin{bmatrix}
    w& & \\
    &\ddots& \\
    & &w  \end{bmatrix} \sigma(\*x) + \begin{bmatrix}b\\\vdots\\ b\end{bmatrix}
\end{equation}
for $\*x\in \R^{d_x}$ and $u=(w,b)\in\R^{d_u}$, with $d_u:=d^2+d$. 
Once again, $\sigma\in \Lip(\R)$ is defined componentwise, so that each component of $\*f$ coincides with the neural ODE given in \eqref{eq: 1.2}. 
Permutations may also be considered, such as
\begin{equation} \label{dyn.sigma.outside intro}
\*f(\*x, u) = \sigma\left(\begin{bmatrix}
    w& & \\
    &\ddots& \\
    & &w  \end{bmatrix} \*x + \begin{bmatrix}b\\\vdots\\ b\end{bmatrix}\right),
\end{equation}
as in the original paper \citep{weinan2017proposal}. 
The key assumption we shall henceforth make regarding $\*f$ is the following.  

\begin{assumption}[Homogeneous dynamics]
We suppose that $\sigma\in\mathrm{Lip}(\mathbb{R})$. We suppose that $\*f$ is $1$--homogeneous with respect to the controls $u$, in the sense that
\begin{equation*} \label{eq: f.homogeneous}
\*f(\*x, \alpha u) = \alpha\,\*f(\*x,u)
\end{equation*}
for all $(\*x,u)\in \R^{d_x}\times\R^{d_u}$ and for all $\alpha>0$.
\end{assumption}

This is clearly the case for dynamics $\*f$ parametrized as in \eqref{dyn.sigma.inside intro}, whilst for \eqref{dyn.sigma.outside intro}, we shall moreover assume that $\sigma$ is $1$--homogeneous -- a prototypical example is the ReLU $\sigma(x) = \max\{x,0\}$, or more general variants such as $\sigma(x)=\max\{ax, x\}$ for $a\in[0,1)$. 
(Such homogeneity assumptions are not an oddity in theoretical contexts, see \citep{chizat2018global} for instance.) Now, as seen in \eqref{eq: 1.4}, given $T>0$ we shall consider the following minimization problem
\begin{equation} \label{functional intro}
\inf_{\substack{u\in\mathfrak{U}_{\text{ad},T}\\ \*x\text{ solves } \eqref{dyn.general}}}\underbrace{\int_0^T \mathscr{E}(\*x(t)) \diff t + \int_0^T \|u(t)\|_1 \diff t}_{:=\mathscr{J}_T(u)},
\end{equation}
where $\mathscr{E}$ is defined in \eqref{empiric loss}, and
\begin{equation*}
\mathfrak{U}_{\mathrm{ad},T} := \Big\{
u\in L^1 (0,T;\R^{d_u}) \colon
\ \| u(t)\|_1 \leqslant M \ \text{ a.e. in } (0,T)\Big\}
\end{equation*}
for a fixed thresholding constant $M>0$. Note that for such controls, \eqref{dyn.general} admits a unique solution $\*x \in C^0([0,T]; \R^{d_x})$ by the Cauchy-Lipschitz theorem.
We postpone commenting the need of having an $L^\infty$ constraint in $\mathfrak{U}_{\text{ad},T}$ to \Cref{rem: constraint}. Before doing so, we make precise the exact assumptions we shall henceforth make regarding the loss function inducing the error $\mathscr{E}$, defined in \eqref{empiric loss}, appearing in \eqref{functional intro}.

\begin{assumption}[The loss function]
We suppose that $\loss(\cdot, \cdot): \R^m\times \mathcal{Y}\to\R_+$ appearing in \eqref{empiric loss} satisfies
\begin{equation*}\label{hyp loss}
\loss(\cdot, y) \in \Lip_{\mathrm{loc}}(\R^m; \R_+) \hspace{0.5cm} \text{ and }
\hspace{0.5cm} \inf_{x\in\R^{m}} \loss(x,y) = 0
\end{equation*}
for all $y\in\mathcal{Y}$.
\end{assumption}

This assumption is generic among most losses considered in practice, including all those induced by a distance (e.g., least squares error) and the cross-entropy loss \eqref{eq: cross.entropy}.

\subsection{Main result} 
Throughout the paper, we will assume that the neural ODE can interpolate the dataset defined in \eqref{eq: dataset}, either in finite or in infinite time. This is an exact controllability assumption, as we shall suppose that there exist controls for which the corresponding stacked trajectory $\*x(t)$ makes $\mathscr{E}(\*x(\cdot))$ (defined in \eqref{empiric loss}) vanish in finite or in infinite time respectively.

\begin{definition}[Interpolation] 
\label{def:approximation}
We say that 
\begin{enumerate}
\item \eqref{dyn.general} \emph{interpolates} the dataset \eqref{eq: dataset} in some time $T>0$ if there exists $T>0$ and $u\in L^\infty(0,T; \R^{d_u})$ such that the solution $\*x\in C^0([0,T]; \R^{d_x})$ to \eqref{dyn.general} satisfies
\begin{equation*}
\mathscr{E}(\*x(T))=0.
\end{equation*} 

\item \eqref{dyn.general} \emph{asymptotically interpolates} the dataset \eqref{eq: dataset} if there exist $T>0$, some function $h\in C^\infty([T,+\infty);\R_+)$ satisfying
\begin{equation*}
\dot{h}<0 \hspace{0.75cm} \text{ and } \hspace{0.75cm} \lim_{t\to+\infty} h(t) = 0,
\end{equation*}
and some $u\in L^\infty(\R_+; \R^{d_u})$
such that the solution $\*x \in C^0(\R_+; \R^{d_x})$ to \eqref{dyn.general} set on $\R_+$ satisfies
\begin{equation*}
\mathscr{E}(\*x(t)) \leqslant h(t) 
\end{equation*}
for $t\geqslant T$.
\end{enumerate}
\end{definition}

These conditions actually hold for the dynamics $\*f$ and many of the errors $\mathscr{E}$ we consider here -- we postpone this discussion to \Cref{rem: asymptotic.interpolation}. We may now state our main result. 

	\begin{theorem} \label{thm: L1.regu}
	Suppose $T>0$ and $M>0$ are fixed. 
	Let $u_T \in \mathfrak{U}_{\mathrm{ad},T}$ be any (should it exist\footnote{One can show that a minimizer exists when $\*f$ is as in \eqref{dyn.sigma.inside intro} by means of the direct method in the calculus of variations. However, for $\*f$ as in \eqref{dyn.sigma.outside intro}, it's not clear if there is enough compactness to convert weak convergences into pointwise ones for passing to the limit inside $\sigma$.}) minimizer of \eqref{functional intro}. 
	Let $\*x_T \in C^0([0,T]; \R^{d_x})$ denote the corresponding solution to \eqref{dyn.general}.
	Then, there exists some time $T^\ast\in(0,T]$ such that
 \begin{align} 
 \|u_T(t)\|_1 &= M \hspace{1cm} \text{ for a.e. }  t\in (0,T^\ast),  \nonumber \\
 \|u_T(t)\|_1 &= 0 \hspace{1.15cm}  \text{ for a.e. }  t\in (T^\ast, T)\label{L1 conclusion 1}.
\end{align}
Moreover, $T^*$ is such that
   \begin{equation}\label{L1 conclusion 2}
   \mathscr{E}(\*x_T(T^\ast)) \leqslant \mathscr{E}(\*x_T (t)) \hspace{1cm} \text{ for }  t\in [0,T],
   \end{equation}
and, furthermore,
\begin{enumerate}
\item If system \eqref{dyn.general} interpolates the dataset in some time $T_0>0$ as per \Cref{def:approximation}, then there exists a constant $\mathfrak{C}>0$ independent of both $T$ and $M$, such that
\begin{equation*}
T^\ast \leqslant \mathfrak{C}\left(\frac{1}{M}+\frac{1}{M^2}\right)
\end{equation*}
and
\begin{equation*}
\mathscr{E}(\*x_T(T^\ast))\leqslant\frac{\mathfrak{C}}{T}\left(\frac{1}{M}+1\right).
\end{equation*}
\smallskip

\item If system \eqref{dyn.general} asymptotically interpolates the dataset as per \Cref{def:approximation}, then there exists a constant $\mathfrak{C}(M)>0$ independent of $T$ such that
\begin{equation*}
T^\ast \leqslant \dfrac{\mathfrak{C}(M)}{M} h^{-1} \left(\dfrac{1}{T}\right) + \dfrac{1}{M}
\end{equation*}
and 
\begin{equation*}
\mathscr{E}(\*x_T(T^\ast)) \leqslant\dfrac{\mathfrak{C}(M)}{T} h^{-1} \left(\dfrac{1}{T}\right) + \dfrac{1}{T},
\end{equation*}
where $h^{-1}$ denotes the inverse function of $h$.
\end{enumerate}
\end{theorem}
\smallskip

\textit{Sketch of the proof.}
In the proof of the theorem, which may be found in \Cref{sec: proofs}, the stopping time $T^*>0$ is precisely defined as 
\begin{equation*}
T^* := \min\left\{t\in[0,T]\colon\mathscr{E}(\*x_T(t))=\min_{s\in[0,T]}\mathscr{E}(\*x_T(s))\right\}.
\end{equation*}
This implies \eqref{L1 conclusion 2} by definition. One then shows that the temporal sparsity in equations \eqref{L1 conclusion 1} holds. This is done by a contradiction argument: one supposes that either of both conclusions doesn't hold, and in both cases, constructs auxiliary controls which are strict minimizers for $\mathscr{J}_T$ defined in \eqref{functional intro}. 
This is quite transparent in the case in which $\|u_T(t)\|\neq 0$ for $t\geqslant T^*$, in which case, one can simply use a zero extension of $u_T(t)$ for $t\geqslant T^*$ to conclude.
On the other hand, if $\|u_T(t)\|<M$ for $t\in(0,T^*)$, the construction is more delicate and technical, and makes crucial use of the scaling provided provided by the homogeneous dynamics, and the invariance of the $L^1(0,T;\mathbb{R}^{d_u})$ by this scaling.
The estimates on the stopping time $T^*$ and on the error evaluated at the stopping time can then be obtained by making use of the interpolation assumptions and the mentioned scaling, for constructing suboptimal controls which can be estimated appropriately.
In particular, our arguments do not rely on studying the first-order optimality system, and is specifically tailored to the particular ODEs in question. This allows us to avoid smallness assumptions on the data, and smoothness assumptions on the nonlinearity.

\subsection{Turnpike property} \label{rem: turnpike}
\begin{figure}[h!]
  \begin{minipage}[t]{0.5\textwidth}
  \mbox{}\\[-\baselineskip]
   \caption{For the experiment of \Cref{fig: fig1}, we see that not only the error $\mathscr{E}(\*x_T(t))$ decays (at least polynomially), but the trajectories $\*x_T(t)$ too reach some stationary point which ought to be near $\text{argmin }\mathscr{E}$. (See the discussion just below.)}
\label{fig: fig3.5}
  \end{minipage}\hfill
  \begin{minipage}[t]{0.5\textwidth}
  \vspace{0.75cm}
  \mbox{}\\[-\baselineskip]
  \begin{overpic}[scale=0.55]{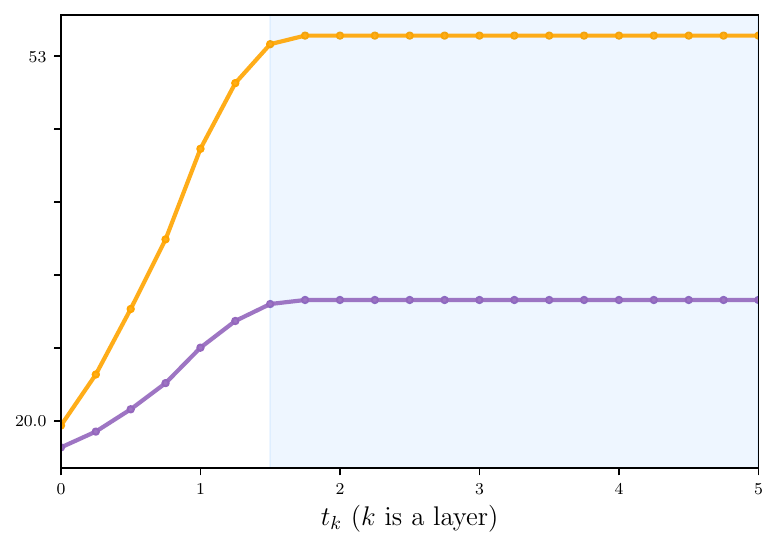}
\put (17, 42) {{\rotatebox{67}{{\color{orange}\tiny$\|\*x_T(t)\|$}}}}
\put (77, 35) {{\rotatebox{0}{{\color{tabpurple}\tiny$\|P\*x_T(t)\|$}}}}
\end{overpic}
  \end{minipage}
\end{figure}

\begin{figure}[h!]

\hspace{-0.35cm}
\begin{tabular}{c}
$t=0$ \\
\includegraphics[scale=0.5]{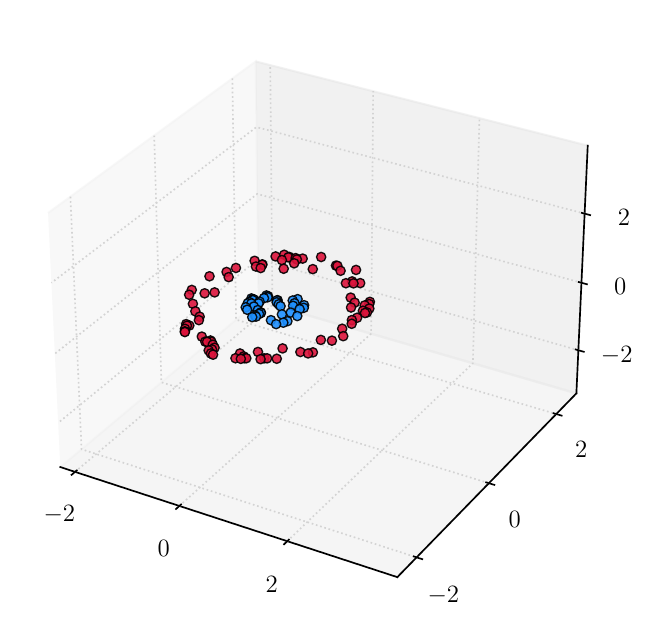}\\
\includegraphics[scale=0.5]{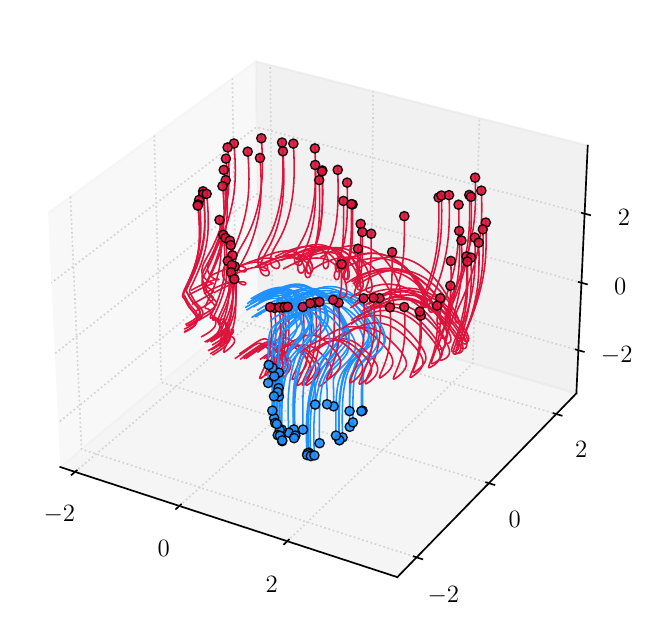}\\
$t=4.25$
\end{tabular}
\hspace{-0.35cm}
\begin{tabular}{c}
$t=0.75$ \\
\includegraphics[scale=0.5]{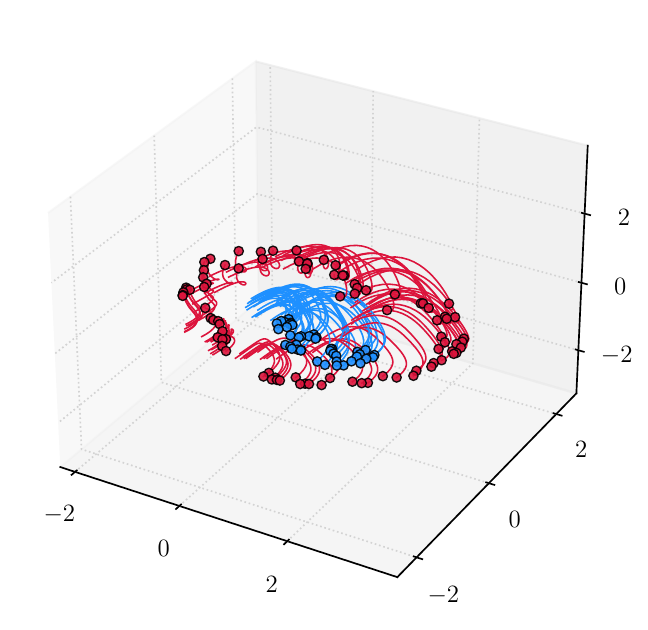}\\
\includegraphics[scale=0.5]{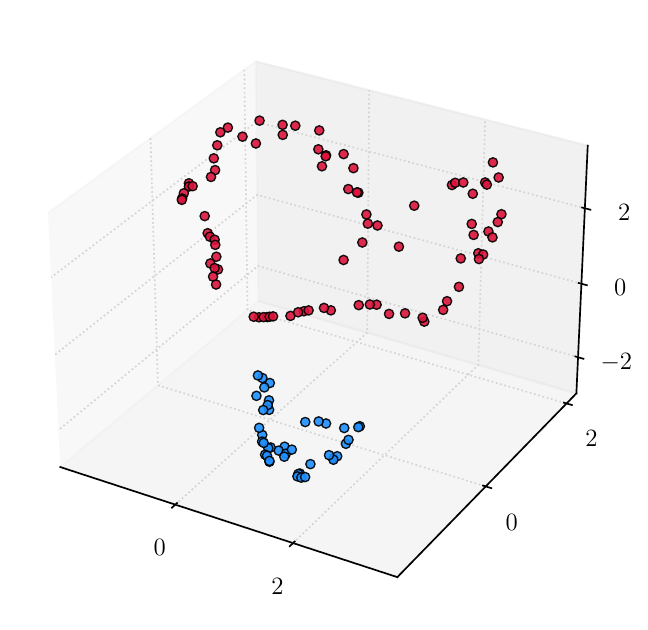}\\
$t=T=5$
\end{tabular}
\caption{The evolution of the states $\{\*x_i(t)\}_{i\in[n]}$ solving \eqref{eq: 1.2}, for the experiment of \Cref{fig: fig1}. The states are stationary in a separation regime beyond $t\geqslant T^*$, as indicated by \Cref{fig: fig1}.}
\label{fig3}
\end{figure}

\begin{figure}
  \begin{minipage}[t]{0.5\textwidth}
   \vspace{0.5cm}
  \mbox{}\\[-\baselineskip]
    \includegraphics[width=\textwidth]{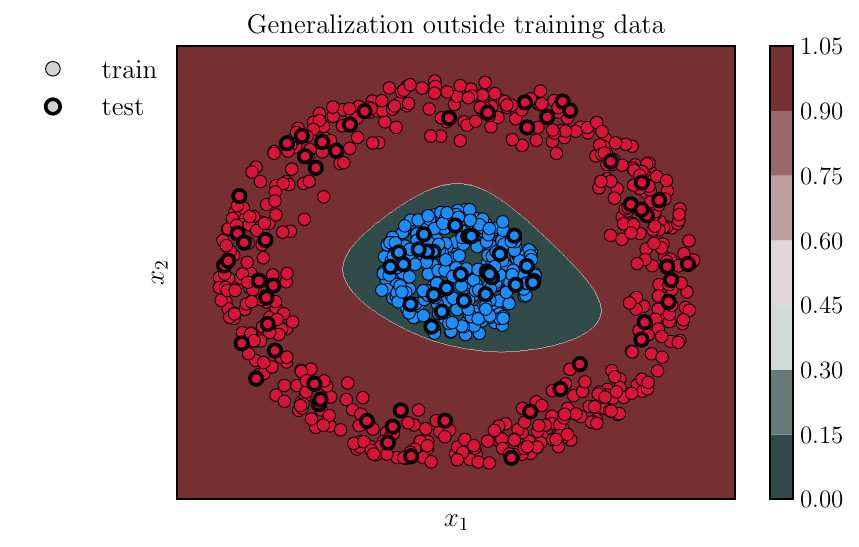}
  \end{minipage}\hfill
  \begin{minipage}[t]{0.5\textwidth}
  \mbox{}\\[-\baselineskip]
    \caption{The predictor $f_{\text{approx}}$ learned through the neural ODE flow. It captures the shape of the dataset given by $f$, accurately classifies the test data, thus ensuring satisfactory generalization.
    } \label{fig4}
  \end{minipage}
\end{figure}

The behavior displayed in \Cref{thm: L1.regu} and \Cref{fig: fig1} -- \Cref{fig4} can, in some contexts, be seen as a novel manifestation of the  \emph{turnpike property} in optimal control: over long time horizons, the optimal pair $(u_T(t), \*x_T(t))$ should be "near" an optimal steady pair $(\overline{u}, \overline{\*x})$, namely a solution to the problem
	\begin{equation} \label{eq: steady.state.ocp}
	\inf_{\substack{(u,\*x) \in \R^{d_u}\times \R^{d_x}\\\*f(\*x,u) =0}} \mathscr{E}(\*x) + \|u\|_1.
	\end{equation}
	(See \citep{geshkovski2022turnpike}.)
	Let us suppose that $\loss(x,y)=\|x-y\|^2_2$ (but the discussion remains true for any distance) and drop the subscript $T$, hence
	\begin{equation*}
	\mathscr{E}(\*x(t)) = \frac{1}{n}\sum_{i=1}^n\left\|P\*x_i(t)-y^{(i)}\right\|^2_2.
	\end{equation*} 
	\Cref{thm: L1.regu} then implies that 
	\begin{equation} \label{eq: turnpike.est}
	\left\|P\*x_i(t)-y^{(i)}\right\|^2_2 \leqslant\frac{C(M)}{T}
	\end{equation}
	for all $t\geqslant T^*$ and $i\in[n]$. Now note that $\*f(\overline{\*x},0)=0$ for any $\overline{\*x}\in\mathbb{R}^{d_x}$. In particular, if $P:\mathbb{R}^d\to\mathbb{R}^m$ is surjective, then taking $\overline{\*x}_i\in P^{-1}\left(\{y^{(i)}\}\right)$ for $i\in[n]$, we see that there exists some $\overline{\*x}\in\mathbb{R}^{d_x}$, with $\overline{\*x}_i\in P^{-1}\left(\{y^{(i)}\}\right)$ such that $(0,\overline{\*x})$ is the unique solution to the steady problem \eqref{eq: steady.state.ocp}. 
	Now, on one hand, the sparsity in time result already ensures a finite-time turnpike property for the optimal controls $u_T(t)$ to the steady correspondent $\overline{u}\equiv0$. 
	On the other hand, \eqref{eq: turnpike.est} can be seen as
	\begin{equation*}
	\Big\|P\Big(\*x_i(t)-\overline{\*x}_i\Big)\Big\|^2_2 \leqslant\frac{C(M)}{T}
	\end{equation*}
	for all $t\geqslant T^*$, $i\in[n]$ and for some $\overline{\*x}_i\in P^{-1}\left(\{y^{(i)}\}\right)$. This is a turnpike property for (a projection of) the state $\*x(t)$. 

Actually, one can see that the above phenomenon is not bound to machine learning, and applies to more classical optimal control problems of the form
	\begin{equation} \label{eq: LQ}
	\inf_{\substack{u\in\mathfrak{U}_{\text{ad},T}\\\*x\text{ solves } \eqref{eq: driftless}}} \int_0^T \|\*x(t)-\overline{\*x}\|_{p}^p + \int_0^T \|u(t)\|_1\diff t,
	\end{equation}
	where $p\in[1,+\infty)$, $\overline{\*x}\in\mathbb{R}^{d_x}$ is fixed, and the 
	underlying system is of \emph{driftless control-affine} form
	\begin{equation} \label{eq: driftless}
	\begin{dcases}
	\dot{\*x}(t) = \sum_{j=1}^{d_u} u_j(t)f_j(\*x(t)) &\text{ in } (0,T),\\
	\*x(0) = \*x^0,
	\end{dcases}
	\end{equation}
	with $f_j:\mathbb{R}^{d_x}\to\mathbb{R}^{d_x}$ for $j\in[d_u]$.
	Then $(\overline{u},\overline{\*x})=(0,\overline{\*x})$ is the optimal steady pair, namely the unique solution to 
	\begin{equation*}
	\inf_{\substack{(u,\*x)\in\mathbb{R}^{d_u}\times\mathbb{R}^{d_x}\\\ \sum_{j=1}^{d_u} u_jf_j(\*x)=0}}\|\*x-\overline{\*x}\|_p^p+\|u\|_1,
	\end{equation*}
	 and we have the following corollary of \Cref{thm: L1.regu}.
	
	\begin{cor}[Turnpike property] 
	\label{cor: turnpike.prop}
	Suppose $\*x_0,\overline{\*x}\in\mathbb{R}^{d_x}$ are given, and let $T>0$, $M>0$ be fixed. Suppose $f_j\in\mathrm{Lip}(\mathbb{R}^{d_x};\mathbb{R}^{d_x})$ for $j\in[d_u]$.
	Let $u_T\in\mathfrak{U}_{\mathrm{ad},T}$ be any solution to \eqref{eq: LQ}. Let $\*x_T$ denote the corresponding solution to \eqref{eq: driftless}. Then there exists some time $T^*\in(0,T]$ and some constant $\mathfrak{C}>0$ independent of both $T$ and $M$ such that
	\begin{equation*}
	 \|u_T(t)\|_1 = M1_{[0,T^*]}(t)
	\end{equation*}
	holds for a.e. $t\in(0,T)$, and
	\begin{equation*}
\|\*x_T(t)-\overline{\*x}\|_p^p \leqslant \frac{\mathfrak{C}}{T}\left(\frac{1}{M}+1\right).
\end{equation*}
holds for all $t\in[T^*,T]$.
	\end{cor}

	\Cref{thm: L1.regu} and \Cref{cor: turnpike.prop} can then be seen as a new result in the turnpike literature: they provide a finite-time, exact turnpike for any optimal control $u_T$ solving \eqref{eq: LQ} (new on its own, due to the $L^1$ penalty of the controls), and a polynomial turnpike for the corresponding optimal state $\*x_T(t)$ for $t\in[T^*,T]$, without any smallness assumptions on the initial data $\*x^0$, on the target $\overline{\*x}$, or smoothness assumptions on the dynamics $f$. The latter are deemed necessary for arguments which make use of the Pontryagin Maximum Principle and linearization (\citep{trelat2015turnpike}). A final arc near $t=T$ doesn't appear as the running cost is at its minimal value for $t\in[T^*,T]$. Another possible approach for proving turnpike would be through the avenue of dissipativity theory in the sense of Willems (see the recent survey \cite{faulwassera2020turnpike}), but due to its non-smooth nature, showing that this problem fits in the dissipativity setting is not straightforward.  
Similar results have been obtained for $L^2$ penalties in \citep{esteve2020large, esteve2022turnpike} (see also \citep{faulwasser2021turnpike, effland2020variational, gugat2020finite, faulwasser2017exact}).

	It is gripping that in \Cref{fig: fig3.5}, we actually see this phenomenon for the trajectories when $\mathscr{E}$ is given by the cross-entropy loss \eqref{eq: cross.entropy}. 
	In this case, $\mathscr{E}$ is not coercive: $\mathscr{E}(\*x(t))$ approaches $0$ only if the margin $\gamma(\*x_T(T))$ defined in \eqref{eq: margin.T0} goes to $+\infty$. Namely, every trajectory $\*x_i(T)$ for $i\in[n]$ ought to grow to $+\infty$ in an appropriate direction in $\mathbb{R}^d$.
	Thus, in this non-coercive case, we do not interpret the graph of \Cref{fig: fig3.5} as a turnpike property, since the turnpike would depend on (and increase with) $T$. Rather, the trajectories $\*x(t)$ become stationary beyond time $t\geqslant T^*$ to some point $\overline{\*x}\in\mathbb{R}^{d_x}$, which is polynomially "sliding" to $+\infty$ (the "argmin" of $\mathscr{E}$) as $T\to+\infty$.
	
\subsection{Discussion}

Let us provide a structured commentary regarding the different assumptions surrounding the above result, possible extensions, and novelty with respect to past literature on both neural ODEs and optimal control.

\begin{remark}[$L^\infty$ constraint] \label{rem: constraint}
Penalizing the $L^1$ norm in \eqref{functional intro} enforces the use of sparse controls, which without an $L^\infty$ constraint, would a priori concentrate near $t=0$ as a Dirac mass. We include the $L^\infty$ constraint in the definition of $\mathfrak{U}_{\mathrm{ad},T}$ in order to prevent such degeneracy. 
One can then recover a Dirac mass centered at $t=0$ when $M\to+\infty$.
\end{remark}


\begin{remark}[Interpolation]
\label{rem: asymptotic.interpolation}
In the case where $\mathscr{E}$ attains its infimum (here $0$), (finite-time) interpolation as per \Cref{def:approximation}, which can be seen as simultaneous or ensemble controllability, has been shown to hold for the dynamics $\*f$ as considered here in several recent works \citep{li2019deep, esteve2020large, agrachev2020control, ruiz2021neural, ruiz2022interpolation, barcena2021optimal, tabuada2022universal}. We have stated it as an assumption in \Cref{thm: L1.regu} to make transparent the ingredients used in the proof.

On another hand, as our setting includes losses which do not attain their infimum, one cannot expect exact interpolation to always hold. This is exemplified by the cross-entropy defined in \eqref{eq: cross.entropy}, which motivates the asymptotic interpolation hypothesis. 
Under the assumption that there exists a control $u\in L^\infty(0,T_0;\mathbb{R}^{d_u})$ for which the \emph{margin} $\gamma=\gamma(\*x(T_0))$ defined as 
\begin{equation} \label{eq: margin.T0}
\gamma(\*x(T_0)):= \min_{i\in [n]} \left\{\Big(P\*x_i(T_0)\Big)_{y^{(i)}} - \max_{\substack{j \in [m]\\ j \neq y^{(i)}}} \Big(P\*x_i(T_0)\Big)_j \right\}
\end{equation}
is positive in some $T_0>0$, in \citep[Proposition 7.4.2]{geshkovski2021control} asymptotic interpolation is shown to hold for the cross-entropy \eqref{eq: cross.entropy} with
\begin{equation*}
h(t) = \log \left(1+ (m-1) e^{-\gamma e^t}\right).
\end{equation*}
\end{remark}


\begin{remark}[The dynamics] \begin{itemize}
\item While there are several works in the literature which prove sparsity in time for controls found by minimizing some functional, even for systems with drifts (unlike ours), the theory is either done for linear systems (\citep{zuazua2010switching, alt2015linear, geshkovski2021optimal}), or nonlinear ones for specific regression functionals and/or differentiable dynamics and/or infinite time horizons (\citep{kalise2017infinite, kalise2020sparse, vossen2006l1}). Similar considerations can be found in the literature on optimal control of multi-agent/mean-field systems (\citep{caponigro2013sparse, fornasier2014mean, caponigro2015sparse}).
 The setting we presented herein makes no such assumptions, and our results can then be seen as complementary to these works. Our consideration of divergences instead of distances in the optimization problem can be seen as a novelty in the optimal control context. 
\smallskip
\item More complicated neural ODEs of the form
\begin{equation} \label{eq: compound.dynamics}
\begin{dcases}
\dot{\*x_i}(t) = w^2(t)\sigma\left(w^1(t)\*x_i(t)\right) &\text{ in } (0,T)\\
\*x_i(0) = \*x^0 
\end{dcases}
\end{equation}
for $i\in[n]$, where $w^2(t)\in\R^{d\times d_{\text{hid}}}$ and $w^1(t)\in\R^{d_{\text{hid}}\times d}$ (we omit the translation control for simplicity), tend to perform well in experiments due to the higher number of controls. 
When $\sigma$ is $1$--homogeneous, and $w^2(t)=\pm1$ or is an orthogonal matrix for all $t$, \Cref{thm: L1.regu} still holds due to the fact that \Cref{lem: scaling} applies for such dynamics. When we remove such assumptions on $w^2(t)$, the technical impediment we encounter is the lack of invariance of the $L^1(0,T; \R^{d_u})$ norm with respect to the natural scaling induced by the equation (\Cref{lem: scaling}). 
Indeed, if one sets $w_1^1(t):=T^{\alpha}w^1(tT)$ and $w_1^2(t):=T^{1-\alpha}w^2(tT)$ for $t\in[0,1]$ and some $\alpha\in(0,1)$, then it can be seen that $\*x^1_i(t):=\*x_i(tT)$ solves \eqref{eq: compound.dynamics} on $[0,1]$. Yet, 
\begin{equation*}
\int_0^T \left\|w^1(t)\right\|_1\diff t + \int_0^T\left\|w^2(t)\right\|_1\diff t = T^{\alpha-1}\int_0^1 \left\|w^1_1(s)\right\|_1\diff s + T^{-\alpha}\int_0^1\left\|w^2_1(s)\right\|_1\diff s.
\end{equation*}
This is incompatible with our proof strategy. 
However, noting the above identity, one could investigate the applicability of our techniques to \eqref{eq: compound.dynamics} and parameter regularizations of the form 
\begin{equation*}
\int_0^T \left\|w^1(t)\right\|_1^{\sfrac{1}{\alpha}}\diff t + \int_0^T\left\|w^2(t)\right\|_1^{\sfrac{1}{1-\alpha}}\diff t,
\end{equation*}
which would be invariant by the above scaling. In such a case, the sparsity pattern should be defined with respect to the regularization one considers. Due to the likely nontrivial nature of the proof, we leave it open.
\end{itemize}
\end{remark}

{\color{black}
\begin{remark}[Dependence on the data] \label{rem: dep.data}
Clearly from \eqref{L1 conclusion 1}, we see that the amplitude of the optimal controls is not the appropriate measure for how these controls depend on the data (unlike the case of $\ell^2$-penalties studied in \cite{esteve2020large}). Similar conclusions apply to the corresponding optimal state, which is stationary (and in the interpolation regime, at least numerically) when the control vanishes. The parameter which does however strongly depend on the data is the stopping time $T^*$, through the constant $\mathfrak{C}$. Looking at the proof of \Cref{thm: L1.regu}, we see that this constant is explicit: 
\begin{equation*}
\mathfrak{C}:=\|u_{T_0}\|_{L^\infty(0,T_0)}\max\left\{1,\int_0^{T_0}\mathscr{E}(\*x_{T_0}(t))\diff t\right\},
\end{equation*}
where $T_0>0$ is arbitrary, and $u_{T_0}$ is any control ensuring controllability in the sense of \Cref{def:approximation}. Such controls typically increase with the euclidean norm of the data in a continuous way, and without smallness assumptions, this dependence may be highly nonlinear. Note that this constant also depends on the ambient dimension $d$ (the \emph{width}), and clarifying the role of $d$ in this context is an open problem.
\end{remark}
}

{\color{black}
\begin{remark}[What about \eqref{empiric loss}?] While we do not demonstrate any long-time pattern for global minima of \eqref{empiric loss}, we may provide a numerical comparison with \eqref{eq: 1.4}, in the setup of \Cref{fig: fig1}. (See \Cref{fig: weak.1}.)
The learned predictor is almost identical to that shown in \Cref{fig3} (albeit learned after $20$ layers), so we omit the plot.

	 \begin{figure}[h]
	 \begin{overpic}[scale=0.5]{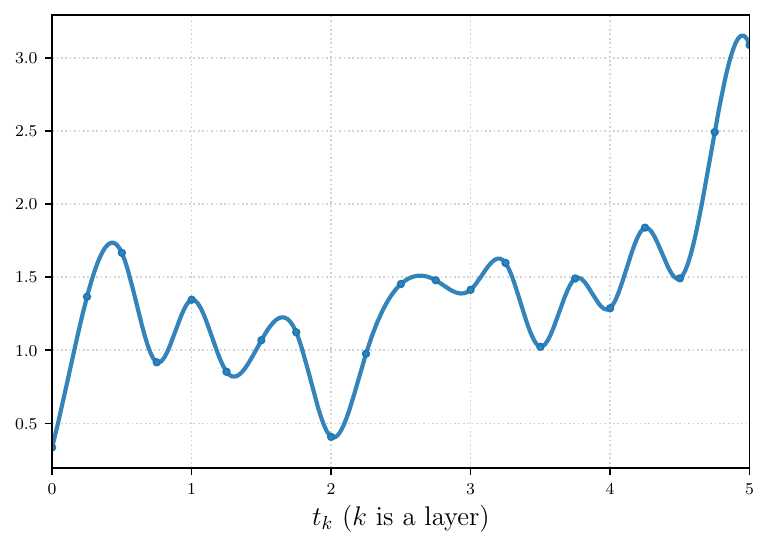}
\put (84.5, 46) {{\rotatebox{75}{{\color{brandeisblue}\tiny$\|u(t)\|_1$}}}}
\end{overpic}
\begin{overpic}[scale=0.5]{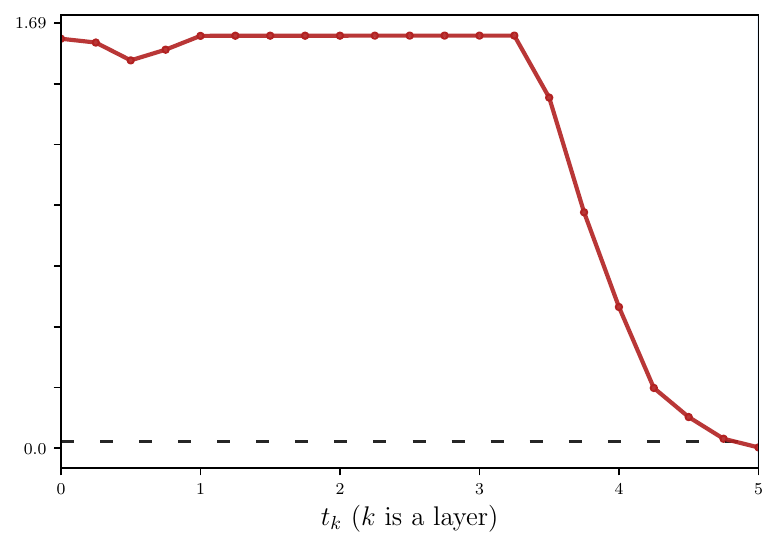}
\put (76, 45) {{\rotatebox{-65}{{\color{firebrick}\tiny$\mathscr{E}(\*x(t))$}}}}
\put (10, 18) {{\rotatebox{0}{\tiny$\mathcal{O}\left(\frac{1}{MT}+\frac{1}{T}\right)$}}}
\end{overpic}
\caption{(\emph{Left}) Optimal controls $u(t)$ solving \eqref{empiric loss}. (\emph{Right}) The empirical risk $\mathscr{E}(\*x(t))$ of the optimal states $\{\*x_i(t)\}_{i\in[n]}$. We do not see any stability, and the empirical risk is small only near the final time/layer.}
\label{fig: weak.1}
\end{figure}


\end{remark}
}

\subsection{Related work} 
The neural ODE lens has been used to great effect in practice. Examples of such use include adaptive ODE solvers \citep{chen2018neural, dupont2019augmented}, symplectic schemes \citep{celledoni2020structure}, or indirect training algorithms based on the Pontryagin Maximum Principle \citep{li2017maximum, benning2019deep}. Further applications include irregular time series modeling \citep{rubanova2019latent, yoon2022learning}, and generative modeling through normalizing flows \citep{grathwohl2018ffjord, papamakarios2021normalizing}.  We refer the reader to the thesis \cite{kidger2022neural} for an excellent review of various applications and state of the art numerical methods.

{\color{black}
Typically in deep learning through neural networks, sparsity is explicitly enforced through the structure of the weights, in the mould of using convolutions with filters instead of matrix multiplications (\cite{mallat2016understanding}). We rather take the approach of considering a somewhat universal architecture (in the spirit of universal approximation setups, \cite{kutyinok}), in view of obtaining a clearer picture on how different penalties affect the long-time properties of global minima (in this regard, see \cite{esteve2020large} for an $\ell^2$-penalty study).
}

\section{Proofs} \label{sec: proofs}

In this section we provide the proof of \Cref{thm: L1.regu}. We shall split the proof into two parts. We first state and prove \Cref{Prop: sparsity}, which contains the first part of \Cref{thm: L1.regu}, concerning the sparsity of optimal controls. The proof of the latter is done throughout \Cref{sec: preliminary.results}. We then provide the remainder of the proof in \Cref{sec: proof.3}.

\subsection{Preliminary results} \label{sec: preliminary.results}

The main goal of this subsection is to state and prove \Cref{Prop: sparsity}. 
A cornerstone of our forthcoming arguments is the possibility of rescaling any trajectory of \eqref{dyn.general} set in $[0,T_0]$ to obtain the same trajectory set on $[0,T]$.
    	
	\begin{lemma} \label{lem: scaling}
	Let $\*x^0 \in \R^{d_x}$, $T_0>0$, $u_{T_0} \in L^1(0,T_0;\R^{d_u})$, and let $\*x_{T_0}$ be the unique solution to \eqref{dyn.general} set on $[0,T_0]$, with control $u_{T_0}$.
	Let $T>0$, and define
	\begin{equation*} \label{eq: uT}
	u_{T}(t):= \frac{T_0}{T} u_{T_0}\left(t\frac{T_0}{T} \right)\hspace{1cm} \text{ for } t\in [0, T],
	\end{equation*}
	and 
	\begin{equation*} \label{eq: xT}
	\*x_{T}(t) := \*x_{T_0}\left(t\frac{T_0}{T}\right) \hspace{1.35cm} \text{ for } t \in [0,T].
	\end{equation*}
	Then $\*x_{T}$ is the unique solution to \eqref{dyn.general} with control $u_{T}$.
	\end{lemma}
	
	We omit the proof, which is straightforward. We also summarize the notion of  sparsity through the following definition.

\begin{definition}[Sparse controls] \label{def:sparse.controls}
Let $M>0$ and $0<T^\ast\leqslant T$ be fixed.
We say that $u\in\mathfrak{U}_{\mathrm{ad},T}$ is \emph{sparse} in $(T^\ast, T)$ if 
\begin{align}
\| u (t)\|_1 &= M \hspace{1.15cm}\text{a.e. }   t\in (0,T^\ast),  \label{sparse.cond.1.1} \\
\| u (t)\|_1 &= 0 \hspace{1.35cm}\text{a.e. }  t\in (T^\ast, T). \label{sparse.cond.1.2}
\end{align}
For any $T^\ast >0$, we shall denote by $\mathfrak{U}_{\mathrm{sp},T^\ast}$ the set consisting of all $u\in\mathfrak{U}_{\mathrm{ad},T}$ which are sparse in $(T^\ast, T)$, namely which satisfy \eqref{sparse.cond.1.1} -- \eqref{sparse.cond.1.2}. 
\end{definition}
	
\begin{proposition}	\label{Prop: sparsity}
Let $T>0$ and $M>0$ be fixed. 
Let $u_T\in\mathfrak{U}_{\mathrm{ad},T}$ be a global minimizer of $\mathscr{J}_T$ defined in \eqref{functional intro}, and let $\*x_T$ be the corresponding unique solution to \eqref{dyn.general}.
Then $u_T\in \mathfrak{U}_{\mathrm{sp},T^\ast}$, where $T^\ast$ is defined as
\begin{equation} \label{eq: def.T*}
T^* := \min \left\{ t\in [0, T] \colon \mathscr{E} (\*x_T (t))= \min_{s\in [0,T]} \mathscr{E} (\*x_T(s)) \right\}.
\end{equation}
\end{proposition}

Note that the $T^*$ is clearly well defined, as the set over which the $\min$ is taken is clearly bounded, and is also closed as the preimage of the singleton 
\begin{equation*}
\displaystyle\left\{\min_{s\in[0,T]} \mathscr{E} (\*x_T (s))\right\}
\end{equation*}
under the continuous map $t \longmapsto  \mathscr{E} (\*x(t))$. 
The core of the proof of \Cref{Prop: sparsity} lies in the following lemma, which ensures that if a control $u_T\in\mathfrak{U}_{\text{ad},T}$ does not saturate the $L^\infty$--constraint before some time $T^\ast$, then $u_T$ is not optimal for $\mathscr{J}_T$ and can always be "improved" through the scaling of \Cref{lem: scaling}.
	
	\begin{lemma} \label{lemma L1 intervals}
Let $T>0$ and $M>0$ be fixed.	
Let $u_T\in \mathfrak{U_{\mathrm{ad},T}}$ be any admissible (but not necessarily optimal) control, and let $T^\ast>0$ be defined as in \eqref{eq: def.T*}.
	Assume that, for some $\theta\in (0,1)$,  there exists a finite collection of disjoint non-empty intervals $\left\{(a_j,b_j)\right\}_{j\in[\mathfrak{I}]}$ with $(a_j, b_j) \subset (0,T^*)$ for which 
\begin{equation} \label{assumption lemma L1 1}
\left\|u_T(t)\right\|_1 \leqslant (1-\theta) M \hspace{2cm} \text{ for a.e. } \ t\in\*O_{\mathfrak{I}},
\end{equation}
and
\begin{equation} \label{assumption lemma L1 2}
\mathscr{E} (\*x_T (t)) - \mathscr{E} (\*x_T (T^\ast)) \geqslant \theta \hspace{0.95cm} \text{ for all } \, t\in\*O_{\mathfrak{I}}
\end{equation}
hold, where
\begin{equation*}
\*O_{\mathfrak{I}}:=\bigcup_{j=1}^{\mathfrak{I}}\left(a_j,b_j\right).
\end{equation*}
Then there exists some $\overline{u}\in\mathfrak{U}_{\mathrm{ad},T}$ satisfying
\begin{equation}\label{conclusion lemma intervals 1}
\overline{u}(t) = 0 \hspace{1cm} \text{ for a.e. } t\in (T^\ast- \tau, T),
\end{equation}
and
\begin{equation*}\label{conclusion lemma intervals 2}
\mathscr{J}_T\left(\overline{u}\right) \leqslant \mathscr{J}_T\left(u_T\right)- \theta \tau,
\end{equation*}
where 
\begin{equation*}
\tau := \theta\,\mathrm{meas}(\*O_{\mathfrak{I}})=\theta \sum_{j=1}^{\mathfrak{I}} (b_j - a_j).
\end{equation*}
\end{lemma}

We may now provide the proof to \Cref{Prop: sparsity}.

\begin{proof}[Proof of \Cref{Prop: sparsity}]

We argue by contradiction. Suppose that $u_T\in \mathfrak{U}_{\mathrm{ad},T}$ is a global minimizer of $\mathscr{J}_T$ such that $u_T\not\in \mathfrak{U}_{\mathrm{sp},T^\ast}$, where $T^\ast>0$ is defined as in the statement. 
Hence, either condition \eqref{sparse.cond.1.1} or condition \eqref{sparse.cond.1.2} does not hold.

\textbf{Case 1: \eqref{sparse.cond.1.2} does not hold.} Let us thus suppose that 
\begin{equation} \label{eq: case1}
\|u_T(t)\|_1>0 \hspace{1cm} \text{ a.e. } t\in \Omega
\end{equation}
holds for some $\Omega\subsetneq(T^*,T)$ of positive Lebesgue measure.
Consider
\begin{equation*}
\overline{u}(t) = \begin{dcases}
u_T(t) &\text{ for } t \in [0,T^*] \\
0 &\text{ for } t \in (T^*, T].
\end{dcases}
\end{equation*}
Clearly $\overline{u}\in\mathfrak{U}_{\mathrm{ad},T}$. Furthermore, we have 
\begin{equation*}
\overline{\*x}(t)=\*x_T(t) \hspace{1cm} \text{ for } t\in[0,T^*],
\end{equation*}
and since $\*f(\cdot,0)\equiv 0$, also
\begin{equation*}
\overline{\*x} (t) = \overline{\*x} (T^\ast) =  \*x_T (T^\ast), \hspace{1cm} \text{ for }  t\in [T^\ast, T].
\end{equation*}
Combining these facts with the definition \eqref{eq: def.T*} of $T^*$, we are lead to
\begin{equation*}
\int_0^T \mathscr{E}(\overline{\*x}(t))\diff t = \int_0^{T^*}\mathscr{E}(\*x_T(t))\diff t +\int_{T^*}^T \mathscr{E}(\*x_T(T^*))\diff t \leqslant \int_0^T \mathscr{E} (\*x_T(t)) \diff t.
\end{equation*}
By virtue of \eqref{eq: case1} we also find
\begin{align*}
\int_0^T \| \overline{u}(t) \|_1 \diff t &= \int_0^{T^\ast} \| u_T(t) \|_1 \diff t \\ 
&<  \int_0^{T^\ast} \| u_T(t) \|_1 \diff t+  \int_{T^\ast}^T \| u_T(t) \|_1 \diff t =  \int_0^T \| u_T(t) \|_1 \diff t.
\end{align*}
Combining the two previous inequalities, we deduce that $\mathscr{J}_T (\overline{u}) < \mathscr{J}_T (u_T)$, which contradicts the optimality of $u_T$.

\textbf{Case 2: \eqref{sparse.cond.1.1} does not hold.} The idea is to again construct an auxiliary control which improves $u_T$ to deduce a contradiction. We now split the proof in three steps.

\textit{Step 1.} If \eqref{sparse.cond.1.1} is not fulfilled, then there must exist some $\theta\in(0,1)$ such that the set
   \begin{equation*}
   \mathbf{A}_\theta := \Big\{t\in (0,T^*) \, \colon \, \left\| u_T(t)\right\|_1 \leqslant (1-\theta) M \Big\}
   \end{equation*}
   has positive Lebesgue measure, namely $\text{meas}(\*A_\theta)>0$.
    Now set $\omega:=\frac{\text{meas}(\*A_\theta)}{2}$, and using elementary set theory we find 
   \begin{equation*}
   \mathbf{A}_\theta\cap(0,T^*-\omega) = \mathbf{A}_\theta \setminus \Big((0,T^*) \setminus(0, T^*-\omega)\Big) = \mathbf{A}_\theta \setminus [T^*-\omega,T^*),
   \end{equation*}
   whence the set
   \begin{equation*}
   \mathbf{B}_\theta := \mathbf{A}_\theta \cap (0,T^*-\omega) 
   \end{equation*}
   also has positive Lebesgue measure: $\text{meas}(\mathbf{B}_\theta)>0$.
    By classical results in Lebesgue measure theory (see \cite[Thm. 3.25]{yeh2006real}), for all $\varepsilon>0$ there exists a finite  collection of disjoint nonempty intervals $\{(a_j,b_j)\}_{j\in[n(\varepsilon)]}$, with $(a_j,b_j)\subset (0,T^*-\omega)$, such that the set
   \begin{equation*}
   \mathbf{O}_\varepsilon:= \bigcup_{j=1}^{n(\varepsilon)} (a_j,b_j)
   \end{equation*}
   satisfies
   \begin{equation}	\label{L1 eps approx interval}
   \text{meas}\left(\mathbf{O}_\varepsilon\setminus \mathbf{B}_\theta \right) <\varepsilon \hspace{0.5cm} \text{and} \hspace{0.5cm}  \text{meas}\left(\mathbf{B}_\theta \setminus \mathbf{O}_\varepsilon\right) <\varepsilon.
   \end{equation} 
   In particular,
   \begin{equation}	\label{proof lemma L1 approx intervals}
   \text{meas}\left(\mathbf{O}_\varepsilon\right) > \text{meas}(\mathbf{B}_\theta) -\varepsilon.
   \end{equation}
   
   \textit{Step 2.} Let $\varepsilon \in (0,\text{meas}(\mathbf{B}_\theta))$ be arbitrary and to be chosen later, and let $\{(a_j,b_j)\}_{j\in[n(\varepsilon)]}$ be the corresponding collection of disjoint intervals satisfying \eqref{L1 eps approx interval}, with $\mathbf{O}_\varepsilon$ denoting the union of these intervals as defined above. We now look to construct a control $u^\varepsilon\in\mathfrak{U}_{\text{ad},T}$ such that 
   \begin{equation*}
   \|u^{\varepsilon}(t)\|_1\leqslant (1-\theta^*)M
   \end{equation*}
   and
     \begin{equation*}
   \mathscr{E}(\*x^\varepsilon(t))-\mathscr{E}(\*x^\varepsilon(T_\bullet))\geqslant\theta^*
   \end{equation*}
   for some $\theta^*>0$ and for all $t\in\*O_\varepsilon$, where
   \begin{equation*}
   T_\bullet : = \min \left\{
   t\in [0,T] \ : \quad \mathscr{E} (\*x^\varepsilon (t)) = \min_{s\in [0,T]} \mathscr{E} (\*x^\varepsilon(s)) 
   \right\}
   \end{equation*}
   should also satisfy $T_\bullet\geqslant T^\ast - \omega$.
  To this end, set
   \begin{equation*}
   u^{\varepsilon}(t) := 
   \begin{dcases}
   u_T(t) &\text{ for }  t\in (0, T) \setminus \left(\mathbf{O}_\varepsilon\setminus \mathbf{B}_\theta \right) \\
    0  & \text{ for } t \in \mathbf{O}_\varepsilon\setminus \mathbf{B}_\theta.
   \end{dcases}
   \end{equation*}
   Since $u_T\in \mathfrak{U}_{\mathrm{ad},T}$, it may readily be seen that
    	\begin{equation*}
        \left\|u^{\varepsilon}(t)\right\|_1\leqslant M \qquad \text{for a.e. $t\in(0,T)$. }
        \end{equation*}
        Hence $u^\varepsilon\in\mathfrak{U}_{\text{ad},T}$.
       Now let $\*x^{\varepsilon}$ denote the solution to \eqref{dyn.general} associated to $u^{\varepsilon}$. By virtue of the  specific form of $\*f$, the Lipschitz continuity of $\sigma$, and the Gr\"onwall inequality, we may readily deduce that there exists a constant $C_1= C_1(T,M,\sigma)>0$ independent of $\varepsilon$ such that
    	\begin{equation}\label{conseq_Gronwall_varepsilon}
    	    \big\|\*x^{\varepsilon}(t)-\*x_T(t)\big\|_1\leqslant C_1\int_0^{T}\big\|u^{\varepsilon}(s)-u_T(s)\big\|_1 \diff s 
    	\end{equation}
	for all $t\in[0,T]$.
    	On the other hand, by using \eqref{L1 eps approx interval}, we also deduce that
        \begin{equation}\label{estimate of uvarespilonTminusuT}
    	    \int_0^T\big\|u^{\varepsilon}(s)-u_T(s)\big\|_1 \diff s \leqslant M \text{meas}\left(\mathbf{O}_\varepsilon\setminus \mathbf{B}_\theta \right) < M\varepsilon.
    	\end{equation}
    	Combining \eqref{conseq_Gronwall_varepsilon} and \eqref{estimate of uvarespilonTminusuT} leads us to
 \begin{equation*}\label{before T'}
\big\|\*x^{\varepsilon}(t)-\*x_T(t)\big\|_1<C_1 M \varepsilon  
\end{equation*}
 for $t\in[0,T]$.
  Now since $\*x_T\in C^0([0,T];\R^{d_x})$, the stacked trajectory $\*x_T(t)$  remains in a compact subset of $\R^{d_x}$ for all $t\in [0,T]$. Due to \eqref{before T'}, and since $\varepsilon\leqslant\text{meas}(\*B_\theta)$, we also find that $\*x_\varepsilon$ remains in a slightly larger compact subset, independent of $\varepsilon$.
  Hence, by the locally Lipschitz character of $\loss\big(\cdot,y\big)$, implying that of $\mathscr{E}$, the estimate
    	\begin{equation}\label{lemma L1 lipsch estimate}
    	    \Big|\mathscr{E}\big(\*x^{\varepsilon}(t)\big) -  \mathscr{E}\big(\*x_T(t)\big)\Big| \leqslant C_2M\varepsilon, \hspace{1cm} 
    	\end{equation}
	holds for some $C_2 = C_2(T, M,\sigma,\mathscr{E})>0$ independent of $\varepsilon$, and for all $t \in [0,T]$. 
	On the other hand, using only the definition \eqref{eq: def.T*} of $T^*$, we find that there exists some $\lambda>0$ such that
   \begin{equation} \label{eq: gamma.eq}
   \mathscr{E}(\*x_T(t))\geqslant \mathscr{E}(\*x_T(T^\ast))+\lambda
   \end{equation}
   for all $t\in[0,T^*-\omega]$. Estimate \eqref{lemma L1 lipsch estimate} combined with \eqref{eq: gamma.eq} yields
   \begin{align}
   \mathscr{E} (\*x^\varepsilon (T^\ast))\leqslant \mathscr{E} (\*x_T(T^\ast)) + C_2 M\varepsilon &\leqslant \mathscr{E}(\*x_T(t)) - \lambda + C_2 M \varepsilon \nonumber \\
   &\leqslant \mathscr{E}(\*x^\varepsilon(t)) - \lambda + 2C_2 M\varepsilon, \label{computation E epsilon}
   \end{align}
   for all $t\in [0,T^\ast-\omega]$, which, by choosing $\varepsilon<\sfrac{\lambda}{2C_2 M}$, implies that 
   $T_\bullet\geqslant T^*-\omega$, as desired.
   The computations done in \eqref{computation E epsilon} also yield   
	\begin{align}
	\mathscr{E}\big(\*x^{\varepsilon}(t)\big)&\geqslant \mathscr{E}\big(\*x^\varepsilon(T^*)\big)  +  \lambda - C_2\,  M \varepsilon\nonumber\\ 
	&\geqslant \mathscr{E}\big(\*x^\varepsilon(T_\bullet)\big) + \lambda - 2C_2 M\varepsilon \label{phi n estimate}
	\end{align}   
	for all $t\in[0,T^*-\omega]$.
As we chose $\varepsilon<\sfrac{\lambda}{2C_2 M}$, we have that $\lambda - 2C_2 M\varepsilon>0$, and may then set
\begin{equation*}
\theta^* := \min\left\{\theta, \lambda-2C_2M\varepsilon \right\},
\end{equation*}
so that $\theta^*>0$. By virtue of \eqref{phi n estimate},
   \begin{equation*}
   \mathscr{E}\big(\*x^{\varepsilon}(t)\big) - \mathscr{E}\big(\*x^\varepsilon(T_\bullet)\big) \geqslant \theta^*
   \end{equation*}
   holds for all $t\in\mathbf{O}_\varepsilon$. Now, observe that $u^{\varepsilon}$ also satisfies
	\begin{equation*}
		 \left\|u^{\varepsilon}(t)\right\|_1 \leqslant \left(1-\theta^*\right) M
	\end{equation*}
	for a.e. $ t\in \mathbf{O}_\varepsilon$. Indeed, if $t\in\*O_\varepsilon\setminus\*B_\theta$, then $u^\varepsilon(t)=0$ by definition, so the inequality clearly holds. On the other hand, if $t\in\*O_\varepsilon\cap\*B_\theta$, then $t\in\*A_\theta$, and since $\theta^*\geqslant\theta$, the conclusion follows.
   
   \textit{Step 3.} We may now apply \Cref{lemma L1 intervals}, which ensures the existence of some $\overline{u}^{\varepsilon}\in\mathfrak{U}_{\text{ad},T}$ for which
   \begin{equation} \label{proof lemma L1 conclusion}
  	\mathscr{J}_T\left(\overline{u}^{\varepsilon}\right) \leqslant \mathscr{J}_T\left(u^{\varepsilon}\right) - (\theta^*)^2 \text{ meas}\left( \mathbf{O}_\varepsilon\right)
   \end{equation}
   holds.
   As a consequence of \eqref{estimate of uvarespilonTminusuT} and \eqref{lemma L1 lipsch estimate}, we have
   \begin{equation*}
  	\mathscr{J}_T\left(u^{\varepsilon}\right) \leqslant \mathscr{J}_T\left(u_T \right) + \left( 1 + C_2T \right) M\varepsilon, 
   \end{equation*}
which, when combined with \eqref{proof lemma L1 conclusion} and \eqref{proof lemma L1 approx intervals},
yields
\begin{equation*}
\mathscr{J}_T\left(\overline{u}^{\varepsilon}\right) < \mathscr{J}_T(u_T) + (1+C_2T)M\varepsilon - (\theta^*)^2 (\text{meas}(\mathbf{B}_\theta) -\varepsilon).
\end{equation*}
Looking at the above inequality, we may note that, by choosing $\varepsilon>0$ even smaller (namely taking
\begin{equation*}
\varepsilon\leqslant \frac{(\theta^*)^2 \text{ meas}(\mathbf{B}_\theta)}{(1+C_2T)M}
\end{equation*}
would do), we may ensure that
\begin{equation*}
\mathscr{J}_T(\overline{u}^{\varepsilon})<\mathscr{J}_T(u_T),
\end{equation*}
which contradicts the optimality of $u_T$.
This concludes the proof.
\end{proof}

We conclude this section with a proof of \Cref{lemma L1 intervals}.

\begin{proof}[Proof of \Cref{lemma L1 intervals}] 
We will argue by induction over the number of intervals $\mathfrak{I} \geqslant 1$, constructing appropriately the control $\overline{u}$ explicitly in each step via affine transformations of $u_T$ -- the desired estimates will follow by using the time-scaling invariance of the $L^1$--norm of the controls.

\textbf{Step 1). Initialization.}
Let us first assume that $\mathfrak{I}=1$. Consider  
   \begin{equation*}
    \overline{u}(t) := \begin{dcases}
    u_T (t) & \text{ for } \ t\in (0,a_1) \\
    \dfrac{b_1-a_1}{c_1-a_1} u_T\left((t-a_1)\dfrac{b_1-a_1}{c_1-a_1} + a_1\right) & \text{ for } \ t\in [a_1,\, c_1) \\ 
    u_T\left(t + b_1-c_1\right) & \text{ for }  t\in [c_1,\, T^*- (b_1-c_1)),  \\
    0 & \text{ for }  t\in [T^*- (b_1-c_1),\, T),
    \end{dcases}
    \end{equation*}
    where $c_1\in (a_1, b_1)$ is chosen so that 
    \begin{equation*}
    \dfrac{b_1-a_1}{c_1-a_1}(1-\theta) = 1,
    \end{equation*}
    which is equivalent to
    \begin{equation*}
     b_1- c_1 = \theta (b_1-a_1)=:\tau.
    \end{equation*}
    Observe that as a consequence of \eqref{assumption lemma L1 1}, we clearly have $\overline{u}\in\mathfrak{U}_{\text{ad},T}$. 
In addition, by virtue of the choice of $c_1$, and the definition of $\tau$, $\overline{u}(t)$ also satisfies \eqref{conclusion lemma intervals 1}.
    Now, making use of the scaling provided by \Cref{lem: scaling}, and the fact that $\*f(\cdot,0)\equiv 0$, one can check that the state trajectory $\overline{\*x}(t)$ associated to $\overline{u}(t)$ is exactly given by
    \begin{equation*}
    \overline{\*x}(t) = \begin{dcases}
    \*x_T(t) & \text{ for } \ t\in [0,a_1) \\
    \*x_T \left((t-a_1)\dfrac{b_1-a_1}{c_1-a_1} + a_1\right) & \text{ for } \ t\in [a_1,\, c_1) \\ 
    \*x_T\left(t + b_1-c_1\right) & \text{ for } \ t\in [c_1,\, T^*- (b_1-c_1)),  \\
    \*x_T (T^*) & \text{ for } \ t\in [T^*-(b_1-c_1) ,\, T].
    \end{dcases}
    \end{equation*} 
    Moreover, observe that since $\tau:=b_1-c_1$,
\begin{equation}\label{induction step 1}
\mathscr{E} (\overline{\*x} (t)) =
\mathscr{E} (\*x_T (T^\ast )) \hspace{1cm} \text{ for } t\in [T^*-\tau, T].  
\end{equation}    
    Let us now evaluate the functional $\mathscr{J}_T$ along $\overline{u}$. 
    We start by computing the $L^1$ norm of $\overline{u}$:    
    \begin{align}
    \left\|\overline{u}\right\|_{L^1(0,T;\R^{d_u})} &= 
    \int_0^{a_1} \left\|u_T (t)\right\|_1\diff t +
     \int_{c_1}^{T^*-(b_1-c_1)} \left\| u_T (t+b_1-c_1)\right\|_1\diff t \nonumber \\
    &\quad+ \dfrac{b_1-a_1}{c_1-a_1}\int_{a_1}^{c_1} \left\| u_T\left((t-t_1)\dfrac{b_1-a_1}{c_1-a_1}+a_1\right) \right\|_1 \diff t \nonumber \\
    &=\int_0^{b_1}\|u_T(s)\|_1\diff s + \int_{T^*-b_1}^{T^*}\|u_T(s)\|_1\diff s \nonumber
    \\
   &\leqslant \left\| u_T\right\|_{L^1(0,T;\R^{d_u})}. \label{L^1 invariant sclaing}
    \end{align}
    On the other hand, by virtue of \eqref{induction step 1}, \eqref{assumption lemma L1 2}, the definition \eqref{eq: def.T*} of $T^*$, and the same changes of variable used to deduce \eqref{L^1 invariant sclaing}, we find
    \begin{align*}
    \int_0^T \Big(\mathscr{E}(\overline{\*x} (t))  -\mathscr{E}\big(\*x_T (T^\ast)\big)  \Big) \diff t  &= 
    \int_0^{a_1} \Big(\mathscr{E}(\*x_T (t)) - \mathscr{E} \big(\*x_T(T^\ast)\big)\Big) \diff t \\
    &\quad+ \underbrace{\dfrac{c_1-a_1}{b_1-a_1}}_{1-\theta} \int_{a_1}^{b_1} \Big(\mathscr{E}(\*x_T (t)) - \mathscr{E}(\*x_T (T^\ast))\Big) \diff t \\
    &\quad  + \int_{b_1}^{T^\ast}   \Big(\mathscr{E} (\*x_T (t)) - \mathscr{E} (\*x_T (T^\ast)) \Big)\diff t \\
   &\leqslant \int_0^T \Big(\mathscr{E}(\*x_T (t))-\mathscr{E}(\*x_T (T^\ast))\Big)  \diff t - \theta^2  (b_1 -a_1).
    \end{align*}    
    By combining the above inequality with \eqref{L^1 invariant sclaing}, it follows that
    \begin{equation*}
    \mathscr{J}_T\left(\overline{u}\right) \leqslant\mathscr{J}_T\left(u_T\right) - \theta^2 (b_1-a_1).
    \end{equation*}
    The statement of the Lemma thus holds for $\mathfrak{I}=1$. 
  
\textbf{Step 2). Heredity.}
    Let us suppose that, for some $n\geqslant 1$, the statement of the lemma holds whenever $\mathfrak{I}=n$, and let $u_T$ satisfy \eqref{assumption lemma L1 1} and \eqref{assumption lemma L1 2} with $\mathfrak{I}=n+1$.
    Assume without loss of generality that $a_1> a_j$ for all $j\in \{2,\ldots,\mathfrak{I}\}$. Using precisely the same argument as in Step 1, we can construct a control $\overline{u}_1$ satisfying
    \begin{equation*}
    \overline{u}_1(t) = 0 \hspace{1cm} \text{ for a.e. }  t\in (T^\ast-\tau_1, T)
    \end{equation*}
  	with $\tau_1 = \theta (b_1 - a_1)$, and  
    \begin{equation*}
    \mathscr{J}_T(\overline{u}_1) \leqslant \mathscr{J}_T\left(u_T\right) - \theta^2 (b_1-a_1),
    \end{equation*}
    and which is such that $\overline{u}_1(t) = u_T(t)$ for all $t\in (0,t_1)$.
    Now observe that, since $a_1>a_j$ for all $j\geqslant 2$,  and in view of \eqref{induction step 1}, it follows that $\overline{u}_1$ satisfies \eqref{assumption lemma L1 1} and \eqref{assumption lemma L1 2} with $\mathfrak{I}-1= n$ number of intervals and with $T_1^\ast = T^\ast - \tau_1$ instead of $T^\ast$.  
    By the induction hypothesis, we conclude that there exists some control $\overline{u}\in \mathfrak{U}_{\mathrm{ad},T}$ such that
    \begin{equation*}
    \overline{u}(t) = 0 \hspace{1cm} \text{ for a.e. } t\in (T^\ast_1-\tau, T)
    \end{equation*}
  with $\tau  = \theta \sum_{j=2}^{\mathfrak{I}}(b_j-a_j)$, and
    \begin{align*}
     \mathscr{J}_T\left(\overline{u}\right) &\leqslant\mathscr{J}_T(\overline{u}_1) - \theta^2 \sum_{j=2}^{\mathfrak{I}} (b_j-a_j) \leqslant\mathscr{J}_T\left(u_T\right)  - \theta^2 \sum_{j=1}^{\mathfrak{
I}} (b_j-a_j).
   \end{align*} 
   The statement of the Lemma thus also holds for $\mathfrak{I}=n+1$.     This concludes the proof.
\end{proof}

 \subsection{Proof of \Cref{thm: L1.regu}} \label{sec: proof.3} 
 
 \begin{proof}[Proof of \Cref{thm: L1.regu}] 
Properties \eqref{L1 conclusion 1} and \eqref{L1 conclusion 2} for the minimizers of $\mathscr{J}_T$ follow directly from \Cref{Prop: sparsity}.  
Let us give the proof of the statements \emph{(i)} and \emph{(ii)} in \Cref{thm: L1.regu}.

\textbf{Proof of \emph{(i)}.} If the interpolation property holds, then there exist $T_0>0$ and some control $u_{T_0}\in L^\infty(0,T_0; \R^{d_u})$ such that the associated solution $\*x_{T_0} \in C^0([0,T_0];\R^{d_x})$ to \eqref{dyn.general} satisfies $\mathscr{E}(\*x_{T_0}(T_0)) = 0$.
Set
\begin{equation} \label{eq: def.T1}
T_1 := \frac{T_0\,\|u_{T_0}\|_{L^\infty(0,T_0; \R^{d_u})}}{M},
\end{equation}
and consider
\begin{equation*}
u_{T_1}(t) := \frac{M}{\|u_{T_0}\|_{L^\infty(0,T_0; \R^{d_u})}} u_{T_0} \left(t\frac{T_0}{T_1}\right) \hspace{1cm} \text{ for } t \in (0,T_1).
\end{equation*}
Observe that $u_{T_1}\in \mathfrak{U}_{\text{ad}, T_1}$. 
Furthermore, in view of \Cref{lem: scaling}, the associated solution $\*x_{T_1}$ to \eqref{dyn.general}, is given by
\begin{equation*}
\*x_{T_1}(t) = \*x_{T_0}\left(t \frac{T_0}{T_1}\right) \hspace{1cm} \text{ for } t\in(0,T_1),
\end{equation*}
and hence, 
\begin{equation*}
\mathscr{E}(\*x_{T_1}(T_1)) = \mathscr{E}(\*x_{T_0}(T_0))=0.
\end{equation*}
Now for any $T>0$, we define
\begin{equation*}
\overline{u}(t) =
\begin{dcases}
u_{T_1}(t) &\text{ for } t \in (0,T)\cap(0,T_1) \\
0 &\text{ for } t \in (0,T)\setminus (0,T_1). 
\end{dcases}
\end{equation*}
Clearly $\overline{u} \in \mathfrak{U}_{\mathrm{ad},T}$.
By a simple change of variable, and using \eqref{eq: def.T1}, one sees that
 \begin{align}
 \mathscr{J}_T (\overline{u}) &\leqslant \int_0^{T_1} \mathscr{E}(\*x_{T_1}(t)) \diff t + M\, T_1 \nonumber \\
 &=  \dfrac{\|u_{T_0}\|_{L^\infty(0,T_0; \R^{d_u})}}{M}\int_0^{T_0} \mathscr{E} (\*x_{T_0} (t)) \diff t + \|u_{T_0}\|_{L^\infty(0,T_0; \R^{d_u})}\, T_0 \nonumber \\
 &= \dfrac{C_1}{M} + C_2,  \label{ineq (i)}
 \end{align}
 holds, where $C_1>0$ and $C_2>0$ are independent of both $T$ and $M$. 
In view of \eqref{L1 conclusion 1}, any minimizer $u_T$ of $\mathscr{J}_T$ satisfies $u_T\in \mathfrak{U}_{\mathrm{sp},T^\ast}$ for some $T^\ast \in (0,T]$.  Since $\overline{u} \in \mathfrak{U}_{\mathrm{ad},T}$, using \eqref{ineq (i)}, we obtain
\begin{align} \label{eq: est.star}
\mathscr{J}_T(u_T) = \int_0^T \mathscr{E} (\*x_T (t)) \diff t +M\, T^\ast  &\leqslant \mathscr{J}_T(\overline{u}) \leqslant \frac{C_1}{M} + C_2.
\end{align}
Since $\mathscr{E}\geqslant0$, using \eqref{eq: est.star} we deduce that 
\begin{equation*}
T^\ast \leqslant\frac{C_1}{M^2} + \frac{C_2}{M}.
\end{equation*}
Moreover, using \eqref{L1 conclusion 2} in \eqref{eq: est.star}, we also deduce that
\begin{align*}
 T \mathscr{E} (\*x_T (T^\ast)) \leqslant \mathscr{J}_T (u_T) \leqslant\frac{C_1}{M} + C_2.
\end{align*}
The last two estimates imply \emph{(i)} in the statement of \Cref{thm: L1.regu}, as desired. 

 \textbf{Proof of \emph{(ii)}.} If the asymptotic interpolation property holds, then there exist $T_0>0$, a function $h$ as in \Cref{def:approximation}, and some control $u^\infty\in L^\infty (\R_+; \R^{d_u})$ such that the corresponding solution $\*x^\infty$ to \eqref{dyn.general} set on $\mathbb{R}_+$ satisfies
 \begin{equation}\label{petite o}
 \mathscr{E}(\*x^\infty(t)) \leqslant h(t)
 \end{equation}
 for all $t\geqslant T_0$.
Combining this with the continuity of the map $t\longmapsto  \mathscr{E} (\*x^\infty (t))$, we can readily deduce that there exists a constant $C_0>0$ depending only on $T_0>0$ such that
\begin{equation} \label{eq: bound.E}
\mathscr{E} (\*x^\infty (t)) \leqslant C_0
\end{equation}
for all $t\geqslant0$.
Let us henceforth set
\begin{equation*}
\lambda:=\frac{M}{\big\|u^\infty\big\|_{L^\infty(\R_+; \R^{d_u})}}.
\end{equation*}
For any $T_1 >0$, we also define
\begin{equation*}
u_{T_1}(t) = 
\begin{dcases}
\lambda u^\infty(\lambda t) &\text{ for } t \in (0,T_1]\\
0 &\text{ for } t > T_1.
\end{dcases}
\end{equation*}
Observe that, by definition of $\lambda$, one has $u_{T_1}\in \mathfrak{U}_{\mathrm{ad},T}$ for any $T>0$.
By virtue of \Cref{lem: scaling}, the state associated to $u_{T_1}$ is precisely
\begin{equation*}
\*x_{T_1}(t) = 
\begin{dcases}
\*x^\infty \left(\lambda t\right) &\text{ for } \ t\in (0, T_1) \\
\*x^\infty \left(\lambda T_1\right) &\text{ for } t \geqslant T_1.
\end{dcases}
\end{equation*}
Now, by virtue of the definition of $u_{T_1}$, for any $T>0$, we have
\begin{align}
\mathscr{J}_T(u_{T_1}) &\leqslant  \int_0^{T_1} \mathscr{E}\left(\*x^\infty \left(\lambda t\right)\right) \diff t + \max\Big\{0, T - T_1\Big\} \mathscr{E} \Big(\*x^\infty \left(\lambda T_1\right)\Big) + M\, T_1 \nonumber \\
&\leqslant  (C_0+M)\, T_1 + T \, \mathscr{E} \Big(\*x^\infty\left(\lambda T_1\right)\Big). \label{ineq (ii)}
\end{align}
We now distinguish two cases. 
If $T\leqslant\sfrac{1}{h(T_0)}$, then using \eqref{eq: bound.E}, the optimality of $u_T$ as well as the fact that $u_{T_1}\in\mathfrak{U}_{\text{ad},T}$, along with $u_T\in\mathfrak{U}_{\text{sp},T^*}$, and the definition \eqref{eq: def.T*} of $T^*$, through \eqref{ineq (ii)} we find
\begin{equation*}
 T\mathscr{E}(\*x_T(T^*))+MT^*\leqslant (C_0+M)T_1 + \frac{C_0}{h(T_0)},
\end{equation*}
and choosing $T_1=1$ leads us to the conclusion.
Now suppose that $T>\sfrac{1}{h(T_0)}$. By \Cref{def:approximation}, the decreasing function $h$ is a bijection from $(T_0,+\infty)$ onto its range $(0,h(T_0))$, and so
$h^{-1}\left(\sfrac{1}{T}\right)$ is well defined precisely for $T>\sfrac{1}{h(T_0)}$.
We set
\begin{equation*}
T_1 := \frac{1}{\lambda} h^{-1} \left(\dfrac{1}{T}\right).
\end{equation*}
Combining the optimality of $u_T$ with \eqref{ineq (ii)}, and using the fact that $u_T\in \mathfrak{U}_{\mathrm{sp},T^\ast}$, we find
\begin{align}
\mathscr{J}_T (u_T) = M\, T^\ast + \int_0^T \mathscr{E}(\*x_T (t))\diff t &\leqslant \mathscr{J}_T (u_{T_1}) \nonumber\\
&\leqslant 
\mathfrak{C}(M)\, h^{-1}\left(\dfrac{1}{T} \right) + T \,  \mathscr{E}\left( \*x^\infty \left( h^{-1} \left(\dfrac{1}{T}\right)  \right) \right),
\label{eq: last.inequality}
\end{align}
where the constant
\begin{equation*}
\mathfrak{C}(M) := \dfrac{(C_0 + M)}{\lambda}
\end{equation*}
is independent of $T$. Now since $h^{-1}: (0, h(T_0))\to(0,+\infty)$ is non-decreasing, and $T>\sfrac{1}{h(T_0)}$, we have that $h^{-1}(\sfrac{1}{T})\geqslant T_0$. Using this fact, along with \eqref{petite o} in \eqref{eq: last.inequality}, combined with the definition \eqref{eq: def.T*} of $T^*$, allows us to deduce that 
\begin{equation*}
T\mathscr{E} (\*x_T(T^\ast)) +MT^\ast  \leqslant \mathfrak{C}(M) \, h^{-1} \left( \dfrac{1}{T}\right) + 1.
\end{equation*}
The desired statement \emph{(ii)} then follows also for $T>\sfrac{1}{h(T_0)}$. This concludes the proof.
\end{proof}

	\section{Concluding remarks} \label{sec: outlook}
	
	\subsection{Epilogue}
	We have presented a manifestation of an ordered sparsity pattern and approximation/stability properties for supervised learning problems for neural ODEs with $L^1(0,T;\mathbb{R}^{d_u})$ penalties.
	Our main result ensures that any solution $u_T$ to \eqref{functional intro} is sparse in time, in the sense that $u_T\equiv0$ on $(T^*,T)$ for some $T^* \in (0,T]$. Under appropriate controllability assumptions, we also provide estimates on the stopping time $T^\ast$, and on the empirical risk $\mathscr{E}(\*x_T(t))$ for $t\geqslant T^*$. 
	
	\subsection{Outlook} We comment some questions that remain regarding our study. 
	\begin{enumerate}
	\item[\textbf{1.}] The existence of minimizers for \eqref{functional intro}-- \eqref{dyn.sigma.outside intro} remains unclear. It can be ensured if one replaces the $L^1$ penalty by a $\mathrm{BV}$ one, for which compactness of minimizing sequences holds. $\mathrm{BV}$ controls fit in the setting of ordered sparsity, unlike $W^{1,1}$ ones, which are continuous. The $\mathrm{BV}$ norm is also invariant with respect to the scaling of \Cref{lem: scaling}. But a complete extension of our arguments to this case would require further work.
	\smallskip
	\item[\textbf{2.}] It is gripping that, when seen in the classical $L^2$ tracking context (i.e. the loss is the squared $\ell^2$ distance) with an $L^1$ penalty for the controls, \Cref{thm: L1.regu} only provides a polynomial turnpike estimate for the state. This is different to the $L^2$ penalty context, presented in \citep{esteve2020large, esteve2022turnpike}, in which an exponential turnpike/stabilization estimate for the state is shown. There is reason to believe that for more specific loss functions, our stability results can be sharpened. 
	\smallskip
	\item[\textbf{3.}] As a matter of fact, since $u_T(t)=0$ for $t\geqslant T^*$, and our numerical experiments show that the state is stable in a regime in which the error $\mathscr{E}$ is $0$, one could also stipulate that a result of the mould $\mathscr{E}(\*x_T(t))=0$ for $t\geqslant T^*$ holds. Such an exact turnpike property for the state has been obtained in the linear setting in \citep{gugat2020finite}. However, the transfer of the techniques of the latter paper to our setting does not appear straightforward. 
	\end{enumerate}
	
	\noindent
	\textbf{Acknowledgments.} We thank Dario Pighin and Enrique Zuazua for insightful discussions.
	\smallskip
	
	\noindent
	{\small{\textbf{Funding:}}}
		{\small{This project has received funding from the European Union's Horizon 2020 research and innovation programme under the Marie Sklodowska-Curie grant agreement No.765579-ConFlex and from the European Research Council (ERC) under the European Union’s Horizon 2020 research and innovation programme (grant agreement NO. 694126-DyCon).}}

	\bibliographystyle{acm}
	\bibliography{refs}{}
	
	\bigskip

\begin{minipage}[t]{.5\textwidth}
{\bf Carlos Esteve-Yag\"ue}\par
Department of Applied Mathematics\par
\hspace{2cm} and Theoretical Physics\par
University of Cambridge\par
Cambridge\par 
CB3 0WA, UK\par
e-mail: \href{mailto:ce423@cam.ac.uk}{\textcolor{dukeblue}{\texttt{ce423@cam.ac.uk}}}
\end{minipage}
\begin{minipage}[t]{.5\textwidth}
  {\bf Borjan Geshkovski}\par
  Department of Mathematics\par
  Massachusetts Institute of Technology\par
  Simons Building, Room 246C\par
  77 Massachusetts Avenue\par
  Cambridge\par
  MA\par
  02139-4307 USA\par
  e-mail: \href{mailto:borjan@mit.edu}{\textcolor{dukeblue}{\texttt{borjan@mit.edu}}}
\end{minipage}

\end{document}